\documentclass[lettersize,journal]{IEEEtran}
\usepackage{amsmath,amsfonts}
\usepackage{algorithmic}
\usepackage{algorithm}
\usepackage{array}
\usepackage[caption=false,font=normalsize,labelfont=sf,textfont=sf]{subfig}
\usepackage{textcomp}
\usepackage{stfloats}
\usepackage{url}
\usepackage{verbatim}
\usepackage{graphicx}
\usepackage{cite}
\usepackage{hyperref}
\hypersetup{hidelinks}

\usepackage{multirow}
\usepackage{cleveref}
\usepackage{color}
\usepackage{array}

\usepackage{amsmath, bm}

\begin{document}

\title{CausalFSFG: Rethinking Few-Shot Fine-Grained Visual Categorization from Causal Perspective}

\author{Zhiwen Yang, Jinglin Xu and Yuxin Peng
        % <-this % stops a space
\thanks{This work was supported by the grants from the National Natural Science Foundation of China (62525201, 62132001, 62432001) and Beijing Natural Science Foundation (L247006, L257005).}% <-this % stops a space
\thanks{Zhiwen Yang and Yuxin Peng are with the Wangxuan Institute of Computer Technology, Peking University, Beijing, 100871, China.}% <-this % stops a space
\thanks{Jinglin Xu is with the School of Intelligence Science and Technology, University of Science and Technology Beijing, Beijing 100083, China.}% <-this % stops a space
\thanks{Corresponding author: Yuxin Peng (e-mail: pengyuxin@pku.edu.cn).}}

\markboth{Journal of \LaTeX\ Class Files,~Vol.~14, No.~8, August~2021}%
{Shell \MakeLowercase{\textit{et al.}}: A Sample Article Using IEEEtran.cls for IEEE Journals}


\maketitle

\begin{abstract}
Few-shot fine-grained visual categorization (FS-FGVC) focuses on identifying various subcategories within a common superclass given just one or few support examples. Most existing methods aim to boost classification accuracy by enriching the extracted features with discriminative part-level details. However, they often overlook the fact that the set of support samples acts as a confounding variable, which hampers the FS-FGVC performance by introducing biased data distribution and misguiding the extraction of discriminative features.
To address this issue, we propose a new causal FS-FGVC (CausalFSFG) approach inspired by causal inference for addressing biased data distributions through causal intervention. Specifically, based on the structural causal model (SCM), we argue that FS-FGVC infers the subcategories (i.e., effect) from the inputs (i.e., cause), whereas both the few-shot condition disturbance and the inherent fine-grained nature (i.e., large intra-class variance and small inter-class variance) lead to unobservable variables that bring spurious correlations, compromising the final classification performance.
To further eliminate the spurious correlations, our CausalFSFG approach incorporates two key components: (1) Interventional multi-scale encoder (IMSE) conducts sample-level interventions, (2) Interventional masked feature reconstruction (IMFR) conducts feature-level interventions, which together reveal real causalities from inputs to subcategories.
Extensive experiments and thorough analyses on the widely-used public datasets, including CUB-200-2011, Stanford Dogs, and Stanford Cars, demonstrate that our CausalFSFG achieves new state-of-the-art performance.
The code is available at \textcolor{blue}{\href{https://github.com/PKU-ICST-MIPL/CausalFSFG_TMM}{https://github.com/PKU-ICST-MIPL/CausalFSFG\_TMM}}.
\end{abstract}


\section{Introduction}
\label{introduction}

\begin{figure*}[t]
  \centering
  \includegraphics[width=0.99\textwidth]{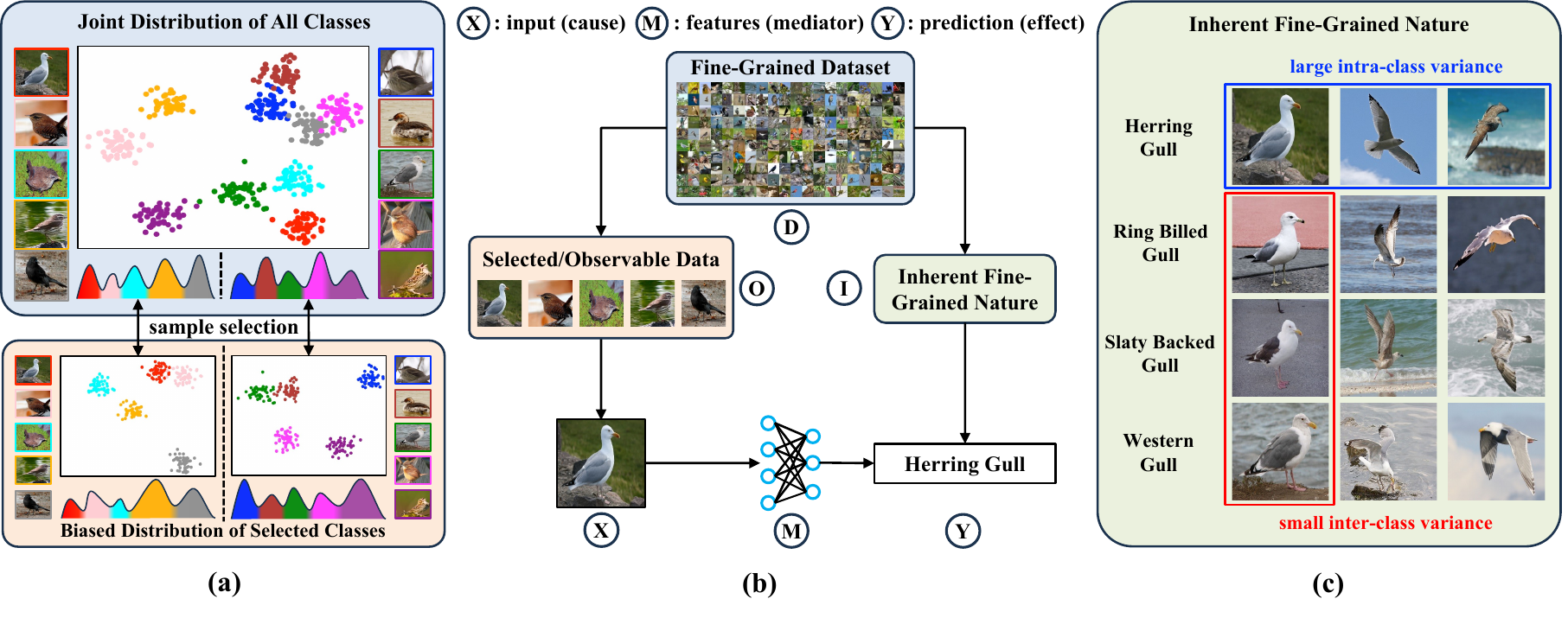}
  \vspace{-15pt}
  \caption{(a) Illustration of the distribution bias caused by the selection operation under the few-shot condition. (b) Illustration of aligning the FS-FGVC problem with the Structural Causal Model assumption, aiming to infer the accurate prediction (effect) from the input (cause) through extracted features (mediator). (c) Illustration of the inherent fine-grained nature of the fine-grained data.}
  \vspace{-7pt}
  \label{motivation}
\end{figure*}

Fine-grained visual categorization (FGVC) is a lasting and crucial problem in computer vision \cite{wei2021fine}, which endeavors to discern various subcategories that belong to a shared superclass. The FGVC task poses significant difficulties stemming from the inherent fine-grained nature of large intra-class variance and small inter-class variance.
Since deep learning has demonstrated great potential in computer vision tasks, existing fine-grained visual categorization techniques leverage deep learning models extensively, which heavily rely on a substantial amount of labeled images
\cite{wang2023learning,liu2023transifc,sun2023hcl,zhang2022fine,sun2022sim,he2019and,peng2017object}. However, different from the traditional image classification task, the labels of fine-grained images in the FGVC task are impractical and expensive due to the scarcity of available samples and the requirement for fine-grained expertise. Therefore, achieving accurate categorization with minimal labeled samples remains challenging.
\par 

In order to emulate the human capacity of acquiring novel knowledge from a limited set of samples \cite{lake2015human}, few-shot fine-grained visual categorization (FS-FGVC) \cite{vinyals2016matching,snell2017prototypical,he2018only} has garnered considerable attention from academic researchers, whose primary objective is to discern novel subcategories given limited support samples. The key challenges of FS-FGVC are twofold: (1) The inherent fine-grained nature, 
where images within the same subcategory can exhibit significant variations and different subcategories often share a high degree of visual similarity.
(2) The few-shot condition, the selected images cannot comprehensively represent the entire subcategory’s variance, even introducing spurious correlations.
To tackle these challenges, existing methods 
adopt several techniques, such as meta-learning~\cite{wertheimer2021few, lee2022task, wu2023bi}, data augmentation~\cite{huang2021attributes, luo2021few, wang2021fine}, transfer learning~\cite{sun2023t2l, tam2025transfer, li2023attention}, and metric learning~\cite{zhang2020deepemd, wu2021selective, liu2023bilaterally}.
These methods primarily emphasize the recognition of discriminative part-level details to address the first challenge for better classification accuracy, but overlook the fact that the set of support samples acts as a confounder, impeding the classification performance with biased data distribution and misguidance in extracting discriminative features.
For instance, when selecting categories exhibiting large variance, such as ``Fish Cow" and ``Yellow Billed Cuckoo", models are prone to focusing on conspicuous coarse-grained features like shapes and outlines, while overlooking discriminative fine-grained features like local textures. Therefore, varied class selections lead to distinct feature distributions that deviate significantly from the joint distribution of the entire dataset, as illustrated in Fig. \ref{motivation} (a).

\par  

To address the aforementioned issues, we reformulate FS-FGVC with a structural causal model (SCM), mitigating the biased data distribution from the few-shot condition via causal intervention.
Beginning with a causal interpretation of FS-FGVC, our CausalFSFG approach illustrates the causalities among the input, features, prediction, fine-grained dataset, observable data, and inherent fine-grained nature, as depicted in Fig. \ref{motivation} (b).
In the context of FGVC, the input samples are directly drawn i.i.d. from the fine-grained dataset, allowing the entire fine-grained dataset to be observed during training. Consequently, the classification model is able to learn from the joint distribution of all subcategories in the fine-grained dataset, and thereby address the challenge of inherent fine-grained nature.
While in the FS-FGVC task,  a subset of specific classes and samples are selected from the fine-grained dataset to construct the observable data comprising certain support and query samples, serving as the input for performing few-shot classification.
As a consequence, the selected subset of observable data hinders the causalities from the fine-grained dataset and the inherent fine-grained nature, introducing biased data distribution to the training process. In our SCM assumption, this subset acts as a confounding variable, and the accompanying biased data distribution further makes the set of confounding variables $(D, O, I)$ unobservable, causing spurious correlations that limit the classification performance.

To eliminate the spurious correlations, we propose a new CausalFSFG approach that learns discriminative features concerning the entire fine-grained dataset on two levels. On the sample level, the interventional multi-scale encoder (IMSE) is proposed to mitigate the biased distributions of selected samples. On the feature level, the interventional masked feature reconstruction (IMFR) is proposed to extract more discriminative features across the query sets. The above complementary modules collectively conduct the causal intervention on the input samples (cause) and the extracted features (mediator), thereby revealing the real causalities from inputs to subcategories for better FS-FGVC performance.

The main contributions can be summarized as follows:

\begin{itemize}
\item We reformulate the FS-FGVC task from the causal perspective to alleviate the biased data distribution caused by the few-shot condition. 
\item We propose the CausalFSFG approach that reveals real causalities from inputs to subcategories by conducting the sample-level and feature-level interventions with an interventional multi-scale encoder (IMSE) and an interventional masked feature reconstruction (IMFR) modules.
\item Extensive comparison experiments on the widely-used public datasets, including CUB-200-2011, Stanford Dogs, and Stanford Cars, demonstrate that our CausalFSFG achieves new state-of-the-art.
\end{itemize}

The rest of the paper is organized as follows: Section \ref{related} provides a brief review of related work on fine-grained visual categorization, few-shot visual categorization, and causal inference. Section \ref{problem} states the detailed definition and parameterization of the FS-FGVC problem, as well as the proposed SCM assumption as its causal reformulation. Section \ref{method} describes the implementation of the proposed CausalFSFG approach based on the frontdoor adjustment rule. Section \ref{experiment} presents the comparison experiments, analyses, and ablation studies. Finally, Section \ref{conclusion} concludes the paper.

\section{Related Work}
\label{related}

This section briefly reviews related works about fine-grained visual categorization, few-shot visual categorization, and causal inference.

\subsection{Fine-Grained Visual Categorization}
Recent FGVC methods predominantly focuses on identifying discriminative regions and extracting features for fine-grained visual classification. Among them, some methods locate distinctive semantic parts and build a mid-level representation for precise classifications, such as utilizing multi-level attention to localize multiple discriminative regions and encode their features in the meanwhile \cite{he2018fast}, and adopting reinforcement learning paradigm to determine the location and number of discriminative regions \cite{he2019and}. Some methods target at modeling subtle differences between fine-grained subcategories in an end-to-end feature encoding manner, such as integrating structure and appearance information to enhance fine-grained representation \cite{sun2022sim}, extracting part-level information and enhancing multi-granularity feature representation~\cite{wang2024multi}. Apart from that, some methods leverage external information to aid the fine-grained classification, such as web data \cite{zhang2020web}, and multi-modal data \cite{song2020bi, jiang2024delving, cheng2024multi}. The aforementioned methods depend heavily on large-scale annotated datasets, while in reality, a substantial amount of data is hard to acquire and costly to label. To mitigate this problem, a more challenging few-shot visual categorization setting is proposed where the model is asked to distinguish fine-grained subcategories with only one or few supporting examples.

\subsection{Few-Shot Visual Categorization}
Given the limited availability of support samples, the meta-learning paradigm has gained popularity for few-shot visual classification. In this framework, the training stage comprises multiple $N$-way $K$-shot episodes designed to simulate the testing stage. Within the meta-learning paradigm, few-shot visual classification methods are generally categorized into two mainstreams: optimization-based and metric-based. \par 
\textbf{Optimization-based methods:} The idea of optimization-based methods were initially introduced in MAML \cite{finn2017model}, which aims to acquire an optimal initialization of model parameters that facilitates smooth fine-tuning. In general, optimization-based methods \cite{antoniou2018train, jamal2019task, lifchitz2019dense, sun2025multi, yang2024channel} start with training the model with auxiliary data, followed by  fine-tuning the network with additional supporting data sampled from unseen classes. MetaOptNet \cite{lee2019meta} harnesses the differentiation conditions and dual formulation of convex optimization problems to boost generalization with high-dimensional embeddings. MattML \cite{zhu2020multi} utilizes a multi-attention mechanism for both base and task learners to locate discriminative part-level details. 
C2-Net~\cite{ma2024cross} integrates outputs from multiple layers with channel activation and position matching operations.
However, one drawback of optimization-based methods is their requirement for online training for novel classes. \par 
\textbf{Metric-based methods:} Metric-based methods \cite{yoon2019tapnet, yang2020dpgn, guo2020attentive, wang2024unbiased, liu2024robust, zhao2024angular, wu2024bi} embed both support and query images into a vector space, and perform classification by distance or similarity metrics. ProtoNet \cite{snell2017prototypical} calculates the average embedding vector of support images as the prototype of each class, and the classification is carried out through distances between a query image and prototypes. RelationNet \cite{sung2018learning} builds a network to learn the suitable distance metric instead of relying on predefined metrics. BSNet \cite{li2020bsnet} develops a bi-similarity network to merge two similarity metrics for learning fewer but more discriminative regions. DUAL ATT-Net \cite{xu2022dual} leverages dual attention streams to model relations among object parts and capture discriminative details. FRN \cite{wertheimer2021few} constructs a feature map reconstruction network that directly regresses from support features to query features in a closed form.
TDM \cite{lee2022task} designs a task discrepancy maximization module for leveraging class-wise channel importance for improved classification. Bi-FRN \cite{wu2023bi} proposes a bi-directional feature reconstruction framework among support and query samples to address both inter-class and intra-class variance. 
BTG-Net~\cite{ma2024bi} filters noise on mid-level features and retains cross-task general knowledge through prompting mechanisms. ATR-Net~\cite{yu2024adaptive} adaptively selects task-specific information by interacting with local feature patches for better integration of task-level and instance-level information.
The above methods primarily concentrate on enhancing the features of selected samples, but are constrained by the biased distributions inherent in this subset. Therefore, we attempt to rectify the confusion arising from the biased distributions from a causal perspective.
\par 

\subsection{Causal Inference}
In the realm of statistics and data science, causal inference serves ad the foundation of uncovering the cause-and-effect relationships underlying observed phenomena. At its essence, causal inference targets at elucidating the mechanisms concerning correlated variables and discerning real causalities from spurious correlations.
Causal inference made its initial attempt into machine learning through the works of \cite{magliacane2018domain, bengio2019meta} and has been adopted in various fields of computer vision since then, such as long-tailed recognition \cite{tang2020long}, semantic segmentation \cite{zhang2020causal}, and image classification \cite{chalupka2014visual}. It is worth noting that the interventional few-shot learning paradigm~\cite{yue2020interventional, wang2025interventional} proposes analyzing the few-shot visual categorization problem with a Structural Causal Model assumption which views the pretrained knowledge as the confounder hindering the classification performance. Despite its effectiveness in the general few-shot classification, IFSL is not compatible with the FS-FGVC problem where the train-from-scratch paradigm eliminates the confounder within the pre-trained knowledge, but instead, the inherent fine-grained nature confuses the subcategory predictions as new confounding variables.\par 

In this paper, we propose a CausalFSFG approach to causally reformulate the FS-FGVC problem with an SCM assumption and address the biased distribution of the limited support samples through the causal intervention.

\section{Problem Reformulations}
\label{problem}

This section states the definition of the FS-FGVC problem and introduces the proposed causal reformulation including the SCM assumption and the casual intervention through the frontdoor adjustment. 

\subsection{Few-Shot Fine-Grained Visual Categorization}
\label{FSFGVC}
In few-shot fine-grained visual categorization scenarios, given a dataset $\mathcal{D}=\{(x_i, y_i), y_i\in C_{total}\}$, we divide it into three subsets: the base training set $\mathcal{D}_{train}=\{(x_i,y_i), y_i\in C_{train}\}$, the validation set $\mathcal{D}_{val}=\{(x_i, y_i), y_i\in C_{val}\}$, and the novel testing set $\mathcal{D}_{test}=\{(x_i, y_i), y_i\in C_{test}\}$, where $x_i$ and $y_i$ denote the $i^{th}$ image and corresponding class label, respectively. The training, validation, and testing classes are disjoint, i.e., $C_{train}\cap C_{val}\cap C_{test}=\phi$. Generally, few-shot visual classification aims to boost an $N$-way $K$-shot classification performance on the testing set $\mathcal{N}$, where $N$ classes are randomly chosen from novel testing classes. Each selected class comprises $K$ labeled images and $U$ unlabelled images. We refer to the labelled images as the support set $S=\{(x_j, y_j)\}_{j=1}^{N\times K}$, and the unlabelled images as the query set $Q=\{(x_j, y_j)\}_{j=1}^{N\times U}$.  \par 
Our approach adopts the meta-learning paradigm, where the training stage is designed to imitate the $N$-way $K$-shot episodes of the testing stage. Concretely, for each episode of the training stage, a meta-training set is randomly sampled from the training set $\mathcal{B}$. Similarly, each meta-training set is composed of $N$ classes from the base train classes $C_{base}$, and each class contains $K$ labeled images and $U$ unlabelled images, forming the support set $S$ and the query set $Q$.

\subsection{Problem Parameterization}
\label{PP}
To further investigate the FS-FGVC problem, we parameterize the problem with the elements depicted in Fig. \ref{motivation} (b): 
\begin{itemize}
  \item $D$: the full fine-grained dataset.
  \item $O$: the observable data for the model during training.
  \item $I$: the inherent fine-grained nature.
  \item $X$: the input samples (cause), which are usually drawn i.i.d. from the observable data $O$.
  \item $Y$: the subcategories (effect) corresponding to $X$.
  \item $M$: the features extracted from models (mediator).
\end{itemize}
First, we can decompose and parameterize the FS-FGVC problem as two successive feature extraction, $P_{\phi}(M)$, and class prediction, $P_{\theta}(Y)$, stages. Specifically, given input samples, the feature maps are extracted conditionally:
\begin{equation}
  \label{feature_extraction}
\begin{aligned}
  P_{\phi}(M)
  = P(M, X)
  = P(M|X)P_{x}(X),
\end{aligned}
\end{equation}
where $P_{x}(X)$ denotes the observed distribution of the input samples. Then, the extracted feature maps are adopted to perform the classification concerning the observed expression of the inherent fine-grained nature:
\begin{equation}
  \label{class_prediction}
\begin{aligned}
  P_{\theta}(Y)
  = P(Y, M, I) 
  = P(Y|M,I)P_{\phi}(M)P_{D}(I),
\end{aligned}
\end{equation}
where $P_{D}(I)$ denotes the expression of the inherent fine-grained nature conditioned in the full fine-grained dataset. Notice that in the FGVC context, the observable data coincides with the full fine-grained dataset, from which the input samples are drawn i.i.d. :
\begin{equation}
  \label{fgvc}
  P_{x}(X) \overset{\text{i.i.d.}}{=}P(D).
\end{equation}
Therefore, by bringing $P_{x}(X)$ into Eqn. (\ref{feature_extraction}) and (\ref{class_prediction}), we can observe that the model has access to the distribution of the full dataset $P(D)$ and thereby the expression of the inherent fine-grained nature $P_{D}(I)$. \par 

While in the FS-FGVC context, the input samples are drawn i.i.d. from the observable data, which is a selected subset of the full fine-grained dataset:
\begin{equation}
  \label{fs_fgvc}
  P_{x}(X) \overset{\text{i.i.d.}}{=}P(O)=P(O|D)P(D).
\end{equation}
It can be observed that the few-shot condition disturbs the observed data distribution with the selection operation $P(O|D)$ which can be decomposed as the class and sample selections:
\begin{equation}
\begin{aligned}
  P(\bm{x},\bm{y}| O)
  & = \sum\limits_{x\in \mathcal{X}}P_{ss}(\mathcal{X}|\mathcal{Y})\sum\limits_{y\in\mathcal{Y}}P_{cs}(\mathcal{Y}|D) \\
  & \neq P(\bm{x},\bm{y}| D),
\end{aligned}
\label{disturb}
\end{equation}
where $x, y$ denotes the input sample and corresponding class label, $P_{cs}$ and $P_{ss}$ denote the class and sample selection operations, $\mathcal{Y}$ is the set of the selected classes, and $\mathcal{X}$ is the set of selected samples. Since the class and sample selections are both random operations, the biased distribution makes the distribution of the full fine-grained dataset together with the inherent fine-grained nature unobservable during training. \par

\subsection{Structural Causal Model}
\label{SCM}
\begin{figure}[!t]
  \centering
  \includegraphics[width=0.47\textwidth]{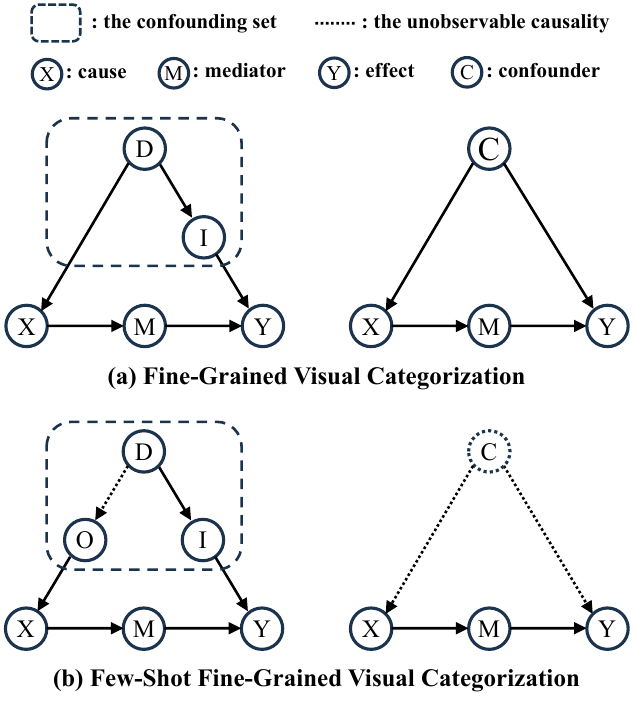}
  \caption{The Structural Causal Model assumptions for the FS-FGVC problem, dashed lines mean the causality is unobservable during training. (a) In the FGVC context, the confounder is observable since the observable data coincides with the full dataset. (b) In the FS-FGVC context, the biased distribution of the selected observable data makes the confounder unobservable.}
  \label{scm}
\end{figure}

From the above discussion, we can see that the FS-FGVC problem can align seamlessly with a Structural Causal Model (SCM), which considers a set of variables associated with the vertices of a directed acyclic graph and depending on their parents in the graph, as illustrated in Fig. \ref{scm}. \par 
\begin{itemize}
  \item $D\rightarrow I$: this link represents that the inherent fine-grained nature is derived from the distribution of the full dataset.
  \item $D\rightarrow O$: this connection represents the causalities from the full fine-grained dataset to the observable data during training. Notice that in the FGVC context, the observable data $D$ coincides with the full fine-grained dataset $D$, as shown in the left part of Fig. \ref{scm} (a). While in the FS-FGVC context, the observable data is a selected subset of the full dataset, as shown in the left part of Fig. \ref{scm} (b).
  \item $O\rightarrow X$: this link represents the causalities that the input samples are drawn i.i.d. from the observable data.
  \item $X \rightarrow M$: this link represents the causalities that the feature maps are extracted depending on the input samples, as formulated in Equ (\ref{feature_extraction}).
  \item $M \rightarrow Y \leftarrow I$: this assumption models the class prediction stage described in Equ (\ref{class_prediction}) and can be interpreted as follows. a) $M \rightarrow Y$: the predicted classification results are directly obtained via the extracted features. b) $I \rightarrow Y$: the inherent fine-grained nature is abstracted from the underlying distribution of the full fine-grained dataset and hence affects the classification implicitly.
\end{itemize} \par 

So far, we have demonstrated the alignment between the SCM assumption and the FS-FGVC problem from the causal perspective. In the following analysis, we refer to the set of the full dataset, observable data, and inherent fine-grained nature as the confounder in the SCM, denoted as $C$, for simplicity. Then, the FS-FGVC problem can be reformulated as follows: 
\begin{itemize}
  \item The main target is to generate correct classification results (effect) corresponding to the input samples (cause) through the extracted features (mediator), while the few-shot condition and inherent fine-grained nature (confounder) together confuse the classification results with spurious correlations.
  \item As formulated in Eqn. \ref{disturb}, in the FS-FGVC problem, the biased distribution is unobservable during training (marked with dashed lines), making the summarized confounder unobservable as well.
\end{itemize}
% The corresponding simplified SCM assumptions are shown in the right part of Fig. \ref{scm}.

\begin{figure*}[!t]
  \centering
  \includegraphics[width=1.0\textwidth]{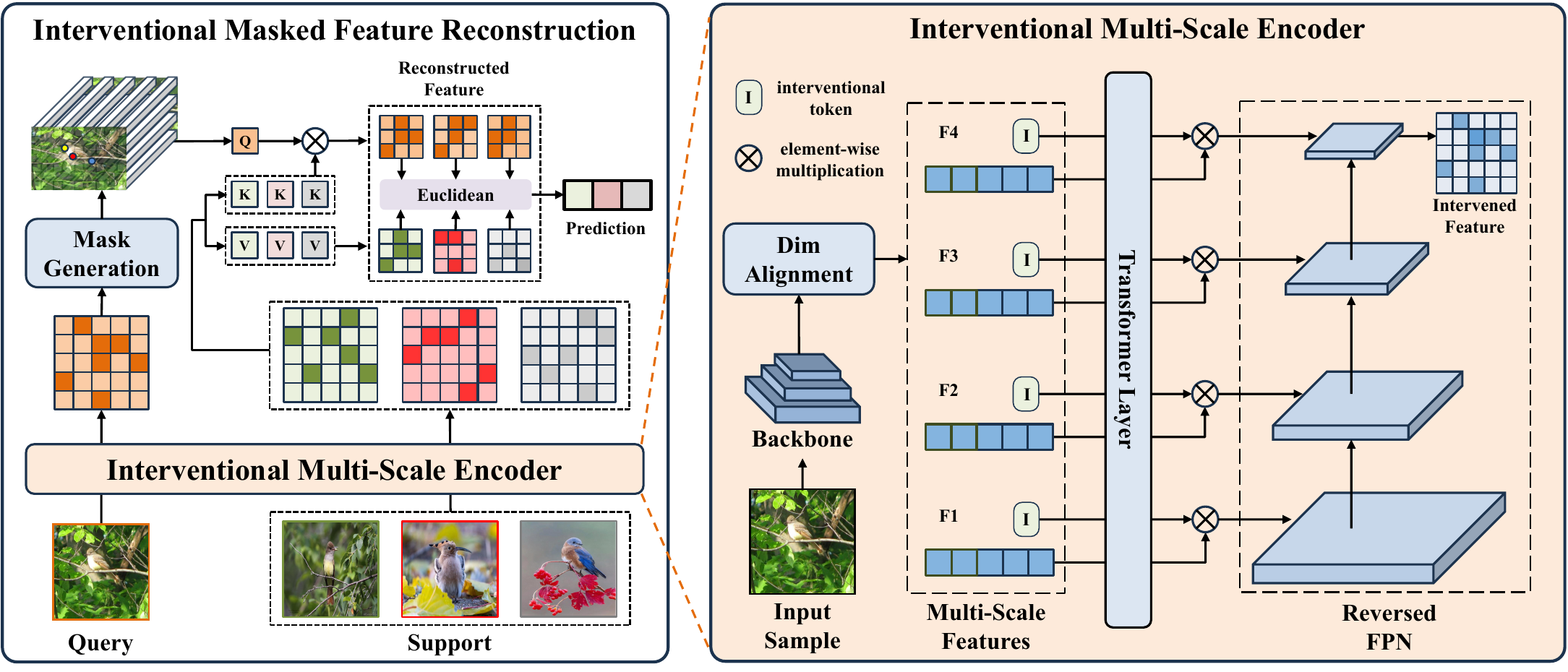}
  \caption{The overall framework of our proposed CausalFSFG approach. The Interventional Multi-Scale Encoder module implements the sample-level intervention to extract the intervened features. Then, the Interventional Masked Feature Reconstruction module further implements the feature-level intervention to improve the classification performance.}
  \label{framework}
\end{figure*}

\subsection{Causal Intervention via Frontdoor Adjustment}
\label{CIFA}
An ideal classification pipeline is to capture the direct causality from $X$ to $Y$, eliminating the confusion from confounder $C$. However, the direct probabilistic likelihood $P(Y|X)$ in the SCM fails to do so since the likelihood $Y$ given $X$ is not only determined by the observable casual path $X\rightarrow M \rightarrow Y$, but also affected by the spurious correlation path $X \leftarrow C \rightarrow Y$. Therefore, to obtain the true causality between input samples $X$ and classification results $Y$, we need to compute the causal intervention \cite{pearl2000models} likelihood $P(Y|do(X))$ instead of the direct likelihood $P(Y|X)$. \par 
Since the confounder $C$ is unobservable under the few-shot condition, which causes the spurious correlations of $X \leftarrow C \rightarrow Y$ cannot be directly computed and eliminated either. To tackle this issue, we adopt the frontdoor adjustment \cite{pearl2016causal} as the causal intervention towards $P(Y|do(X))$. Frontdoor adjustment estimates the causal intervention likelihood $P(T|do(X))$ via intervention on the mediator $M$:
\begin{equation}
\label{frontdoor}
\begin{aligned}
    P(Y|do(X)) 
    &=\sum\limits_{m}P(Y|do(X), M)P(M|do(X))\\
    % &=\sum\limits_{m}P(Y|do(X), do(M))P(M|do(X))\\
    &=\sum\limits_{m}P(Y|do(M))P(M|do(X)) \\
    &=\sum\limits_{m}P(M|X)\sum\limits_{x}P(Y|M,X)P(X)
\end{aligned}
\end{equation}

It can be observed that frontdoor adjustment conducts two intervention operations based on the mediator $M$, i.e., the sample-level intervention $P(M|do(X))$ and the feature-level intervention $P(Y|do(M))$. In the following section, we will present our implementation of the frontdoor adjustment, CausalFSFG, based on the above two intervention operations.

\section{Methodology}
\label{method}
This section introduces the overall pipeline of the proposed CausalFSFG approach and elaborates on each component.

\subsection{Overview}
The whole pipeline of the proposed CausalFSFG approach is illustrated in Fig. \ref{framework}. Inspired by the frontdoor adjustment, our CausalFSFG approach eliminates the spurious correlations caused by the unobserved inherent fine-grained nature (confounder) through the causal intervention. Specifically, an Interventional Multi-Scale Encoder (IMSE) module is proposed as an implementation for $P(M|X)$ in Eqn. (\ref{frontdoor}) and executes the sample-level intervention. Furthermore, an Interventional Masked Feature Reconstruction (IMFR) module is proposed as an implementation for $P(Y|M,X)$ in Eqn. (\ref{frontdoor}) and executes the feature-level intervention.

\subsection{Interventional Multi-Scale Encoder}
We propose to integrate multi-scale features conditionally, based on the discovery that multi-scale features extracted from different convolution layers of the backbone contain complementary information \cite{du2020fine}. \par 
% Our implementation idea is inspired by the property that the multi-scale features extracted from different convolution layers of the backbone contain complementary information \cite{du2020fine}, which should be integrated conditionally. \par 

Suppose that given an input sample $X$, the extracted multi-scale features consist of feature maps of 4 scales (take the Conv-4 and ResNet-12 backbone as examples), i.e., $M=\{M_1, M_2, M_3, M_{4}\}$. In most previous methods, only the last feature map $M_4$ is further processed to derive the classification results. In other words, a one-hot distribution $[0,0,0,1]$ is applied to the extracted multi-scale features, where all feature maps but the last one are overlooked. \par 

To make full use of the multi-scale features and execute sample-level intervention, we implement the target conditional distribution $P(M|X)=P([M_1,M_2,M_3,M_4]|X)$ by introducing interventional tokens which:
\begin{itemize}
  \item promotes the extraction of discriminative features of each scale as general representations,
  \item guides the conditional integration of the extracted multi-scale discriminative features.
\end{itemize}
To begin with, we conduct the dim alignment with $1\times 1$ convolution layers to resize multi-scale features with a unified embedding dim $\Gamma+1$, for each feature map $M_i$:
\begin{equation}
  [F_i, I_i] = DA(M_i),
\end{equation}
where $DA(\cdot)$ denotes the dim alignment operation, $F_i$ is the resized feature map taking up the first $\Gamma$ embedding dim, and $I_i$ is the corresponding interventional token at the last one embedding dim.\par 

Then, the resized multi-scale features and the corresponding interventional tokens are flattened and went through a standard Transformer layer to capture discriminative information of each sale. In the meanwhile, the interventional tokens work as general representations which generalize the discriminative features of each scale into one dimension: 
\begin{equation}
    [F_i, I_i]= \text{Softmax}(\dfrac{Q_{i}K_{i}^{T}}{\sqrt{\Gamma}})V_{i},
\end{equation}
where $Q_{i}, K_{i}, V_{i}$ are a set of learnable weight parameters projected from the feature map and corresponding interventional token of scale $i$.\par 

Notice that the interventional tokens can be considered as the general representations of the multi-scale features, we implement the target distribution $P(M|X)$ by applying the softmax function to the interventional tokens:
\begin{equation}
  [I_{1}^{'},\cdots,I_{4}^{'}] = \text{Softmax}([I_{1},\cdots,I_{4}]).
\end{equation}

So far, we have implemented the target distribution over as $P(M|X)=[I_{1}^{'},\cdots,I_{4}^{'}]$, which are employed to guide the integration of the multi-scale features $[F_{1},\cdots,F_{4}]$. The element-wise multiplication is conducted to get weighted multi-scale features $F_{i}^{'}=F_i \otimes I_{i}^{'}$, which are then integrated by a reversed Feature Pyramid Network (FPN) \cite{lin2017feature}:
\begin{equation}
  F_{I}=F_{4}^{'}+\text{Maxpool}(F_{3}^{'}+\cdots+\text{Maxpool}(F_{1}^{'})),
\end{equation}
where $\mathcal{F}_{I}$ is the output intervened feature.

\subsection{Interventional Masked Feature Reconstruction}
Notice that the following decomposition accords with the few-shot classification paradigm:
\begin{equation}
  \begin{aligned}
      &\sum\limits_{x}P(Y|M, X)P(X) \\
      =& \dfrac{1}{|S|}\sum\limits_{s\in S}P(Y|M,\bm{s})+\dfrac{1}{|Q|}\sum\limits_{q\in Q}P(Y|M,\bm{q})
  \end{aligned}
\end{equation}
where $S$ and $Q$ denote the support and query set respectively. \par 

It can be observed that given an input sample and its intervened feature $M$, the execution of the feature-level intervention requires two implementations:
\begin{itemize}
    \item $P(Y|M,\bm{q})$: This query term indicates the information interaction within the query set, which is implemented with a shared mask generation block for all query samples.
    \item $P(Y|M,\bm{s})$: This support term indicates the information exchange across the query set and the support set, which is implemented with a masked feature reconstruction operation between the query samples and the support samples.
\end{itemize} \par
More concretely, for the $N$-way $K$-shot classification task, we first apply the IMSE module to extract support features of the $n^{th}$ class, i.e. $S_{i}=[\bm{s}_{k}^{n}]\in\mathbb{R}^{C\times H\times W}$, where $n\in[1,2,\cdots, N]$ and $k\in[1,2,\cdots, K]$, and query features $\bm{q}_{i}\in\mathbb{R}^{C\times H\times W}$, where $i\in[1, 2, \cdots, |Q|]$, where $|Q|$ is the number of query samples.\par 

To facilitate the information interaction within the query set, we perform channel-wise max-pooing and average-pooling for each query feature $q_{i}\in C\times H\times W$, integrating global signals as two $1\times H\times W$ tensors. Subsequently, these two tensors are concatenated to form a tensor of size $2\times H\times W$, and a convolution block followed by a sigmoid function is then utilized to produce a normalized global matrix $G_i\in\mathbb{R}^{H\times W}$. \par 

Then we select $k$ elements of $G_i$ with the highest activation values as the indicative mask of general discriminative regions within the query set:
\begin{equation}
  \mathcal{G}_i(h,w)=\left\{
    \begin{array}{ll}
      1, & G_{i}(h,w)\in \text{top-}k(G_i)\\
      0, & \mbox{otherwise}
    \end{array}
  \right.,
\end{equation}
where $\text{top-}k(G_i)$ denotes the set of top-$k$ highest values among the elements of $G_i$. Notice that the convolution block is shared by all query samples, enabling $\mathcal{G}_i$ to implicitly indicate the information interaction within the query set. Therefore, a residual connection is conducted:
\begin{equation}
  \hat{\bm{q}}_i = \bm{q}_i + \bm{q}_i \times \mathcal{G}_i,
\end{equation}
which enhances the query features implements the query term as $P(Y|M,\bm{q})=P(Y|M+M\times\mathcal{G}(Q))$. \par 

Based on the enhanced query features, we develop a direct implementation of the support term $P(Y|M,s)$ with a feature reconstruction operation. The support features of the same class are first averaged into a prototype of the class:
\begin{eqnarray}
  \bm{s}_n = \dfrac{1}{K}\sum\limits_{k=1}^{K}\bm{s}_k^n,\quad n=[1,\cdots,N].
\end{eqnarray}
Then for each prototype $\bm{s}_j$, the feature reconstraction is conducted as follows:
\begin{equation}
    \bm{q}_{i}^{j} = \text{Softmax}(\dfrac{Q_{i}K_{j}^{T}}{\sqrt{\Gamma}})V_{j},
\end{equation}
where $Q_{i}$ is the query matrix projected from $\hat{\bm{q}}_i$ and $K_j, V_j$ are the key and value matrices projected from $\bm{s}_j$. 

\subsection{Learning Objectives}
Regarding the query features $\bm{q}_i$, we compute the Euclidean distance between the reconstructed query feature $\bm{q}_i^j$ and the projected value matrix of support the feature $V_j$:
\begin{equation}
  d_{ij} = ||\bm{q}_i^j-V_j||_2,
\end{equation}
and the predicted probability of the query sample belonging to class $j$ is:
\begin{eqnarray}
  p_{ij} = \dfrac{e^{-d_{ij}}}{\sum_{n=1}^{N}e^{-d_{in}}}.
\end{eqnarray}
The cross-entropy loss is adopted as the final optimization objective to train the model:
\begin{equation}
  \mathcal{L} = CE([p_{i1},\cdots,p_{iN}], \bm{y}_i),
\end{equation}
where $y_i$ is the correct subcategory label.

\section{Experiments}
\label{experiment}

In this section, we assess the efficacy of the proposed CausalFSFG approach on three widely used fine-grained benchmark datasets. We begin by providing a concise overview of these three datasets and how data pre-processing and separation are conducted. Subsequently, we present our implementation details for all experiments. Following this, we compare our approach with state-of-the-art methods to showcase the effectiveness of CauFSFG. Finally, extensive experimental analyses are presented including ablation studies and parameter analysis to validate the contribution of each component in our approach.

\subsection{Datasets and Evaluation Metric}
Three widely used fine-grained datasets, i.e., the CUB-200-2011, Stanford Dogs, and Stanford Cars datasets, are adopted in the experiments. Detailed descriptions of the three datasets are as follows:
\begin{itemize}
  \item CUB-200-2011 (CUB) \cite{wah2011caltech} consists 11,788 images representing 200 bird species. Following \cite{wu2023bi}, the input images are cropped using human-annotated bounding boxes. 
  \item Stanford Dogs (Dogs) \cite{khosla2011novel} comprises 20,580 images across 120 dog breeds.
  \item Stanford Cars (Cars) \cite{krause20133d} includes 16,185 images depicting 120 different car types.
\end{itemize}
For each dataset, we follow \cite{wu2023bi, zhu2020multi} to split the original images into three disjoint subsets: $D_{train}, D_{val}, D_{test}$ for training, validation, and testing respectively. Details are shown in Table \ref{data split}. \par 

We adopt the widely used classification accuracy as the metric to validate the performance of the proposed CausalFSFG approach and other comparison methods.

\subsection{Implementation Details}
For a fair comparison with state-of-the-art methods, we adopt two widely used backbone networks: Conv-4 and ResNet-12. The input images are resized to 84 $\times$ 84, and the output is a feature map with 64 $\times$ 5 $\times$ 5 elements for the Conv-4 backbone and 640 $\times$ 5 $\times$ 5 elements for the ResNet-12 backbone. We apply data augmentation, which includes random crops, random horizontal flips, color jitter at the meta-training stage, and center crops at the testing stage, in all implemented experiments. \par 
As for the meta-learning paradigm, we adopt two different settings for more comprehensive comparisons with the state-of-the-art methods:
\begin{itemize}
  \item Following  \cite{wu2023bi}, we adopt 30-way 5-shot episodes for the Conv-4 backbone and 15-way 5-shot episodes for the ResNet-12 backbone.
  \item Following \cite{lee2022task, li2023locally}, we adopt 5-way 5-shot episodes for both the Conv-4 and ResNet-12 backbones.
\end{itemize}

\begin{table}[t]
  \centering
  \caption{Category split for three datasets. $C_{total}$, $C_{train}$, $C_{val}$, $C_{test}$ represents the number of subcategories in the whole dataset, training set, validation test, and testing set, respectively.}
  \begin{tabular}{|c|c|c|c|}
  \hline
  Dataset & CUB & Cars & Dogs \\
  \hline
  \hline
  $C_{total}$ & 200 & 196 & 120 \\
  \hline
  $C_{train}$ & 130 & 130 & 70 \\
  \hline
  $C_{val}$ & 20 & 17 & 20 \\
  \hline
  $C_{test}$ & 50 & 49 & 30 \\
  \hline
  \end{tabular}
  \label{data split}
\end{table}
\vspace{-2pt}

In both scenarios, we conduct tests for 5-way 1-shot and 5-way 5-shot episodes, selecting 15 query images for each class. The unified embedding dimension $\Gamma$ is configured as 128 for the Conv-4 backbone and 256 for the ResNet-12 backbone. The top-$k$ threshold is designated as $k=5$ for Conv-4 and $k=3$ for ResNet-12. The results are presented as mean accuracy (MA) with 95\% confidence intervals across 10,000 sampled testing episodes. During meta-training, all models are trained from scratch in an end-to-end fashion. Training spans 800 epochs for both Conv-4 and ResNet-12 models, utilizing the SGD optimizer with Nesterov momentum set to 0.9. The initial learning rate is 0.1 with a weight decay of 3e-4. Throughout training, the learning rate diminishes by a scaling factor of 20 every 400 epochs. Experiments are conducted using PyTorch on a single NVIDIA GeForce RTX 4090 GPU.

\begin{table*}
  \begin{center}
  \caption{Comparison results mean$_{\pm \text{std}}$ on the CUB, Dogs, and Cars datasets with the Conv-4 backbone. \textbf{Bold value} indicates the best performance and \underline{underline value} indicates the suboptimal performance.}
  \label{conv4}
  \begin{tabular}{|c|c|c|c|c|c|c|c|}
  \hline
  \multirow{2}*{Method} & \multirow{2}*{Published In} & \multicolumn{2}{c|}{CUB} & \multicolumn{2}{c|}{Dogs} & \multicolumn{2}{c|}{Cars}\\ \cline{3-8}
  & & 1-shot & 5-shot & 1-shot & 5-shot & 1-shot & 5-shot \\
  \hline
  \hline
  ProtoNet \cite{snell2017prototypical} & NeurIPS 2017 & 64.82$_{\pm 0.23}$ & 85.74$_{\pm 0.14}$ & 46.66$_{\pm 0.21}$ & 70.77$_{\pm 0.16}$ & 50.88$_{\pm 0.23}$ & 74.89$_{\pm 0.18}$ \\
  \hline
  Relation \cite{sung2018learning} & CVPR 2018 & 63.94$_{\pm 0.92}$ & 77.87$_{\pm 0.64}$ & 47.35$_{\pm 0.88}$ & 66.20$_{\pm 0.74}$ & 46.04$_{\pm 0.91}$ & 68.52$_{\pm 0.78}$ \\
  \hline
  DN4 \cite{li2019revisiting} & CVPR 2019 & 57.45$_{\pm 0.89}$ & 84.41$_{\pm 0.58}$ & 39.08$_{\pm 0.76}$ & 69.81$_{\pm 0.69}$ & 34.12$_{\pm 0.68}$ & 87.47$_{\pm 0.47}$ \\
  \hline
  PARN \cite{wu2019parn} & ICCV 2019 & 74.43$_{\pm 0.95}$ & 83.11$_{\pm 0.67}$ & 55.86$_{\pm 0.97}$ & 68.06$_{\pm 0.72}$ & 66.01$_{\pm 0.94}$ & 73.74$_{\pm 0.70}$ \\
  \hline
  SAML \cite{hao2019collect} & ICCV 2019 & 65.35$_{\pm 0.65}$ & 78.47$_{\pm 0.41}$ & 45.46$_{\pm 0.36}$ & 59.65$_{\pm 0.51}$ & 61.07$_{\pm 0.47}$ & 88.73$_{\pm 0.49}$ \\
  \hline
  DeepEMD \cite{zhang2020deepemd} & CVPR 2020 & 64.08$_{\pm 0.50}$ & 80.55$_{\pm 0.71}$ & 46.73$_{\pm 0.49}$ & 65.74$_{\pm 0.63}$ & 61.63$_{\pm 0.27}$ & 72.95$_{\pm 0.38}$ \\
  \hline
  LRPABN \cite{huang2020low} & TMM 2021 & 63.63$_{\pm 0.77}$ & 76.06$_{\pm 0.58}$ & 45.72$_{\pm 0.75}$ & 60.94$_{\pm 0.66}$ & 60.28$_{\pm 0.76}$ & 73.29$_{\pm 0.58}$ \\
  \hline
  BSNet(D\&C) \cite{li2020bsnet} & TIP 2021 & 62.84$_{\pm 0.95}$ & 85.39$_{\pm 0.56}$ & 43.42$_{\pm 0.86}$ & 71.90$_{\pm 0.68}$ & 40.89$_{\pm 0.77}$ & 86.88$_{\pm 0.50}$ \\
  \hline
  CTX \cite{doersch2020crosstransformers} & NeurIPS 2020 & 72.61$_{\pm 0.21}$ & 86.23$_{\pm 0.14}$ & 57.86$_{\pm 0.21}$ & 73.59$_{\pm 0.16}$ & 66.35$_{\pm 0.21}$ & 82.25$_{\pm 0.14}$ \\
  \hline
  FRN \cite{wertheimer2021few} & CVPR 2021 & 74.90$_{\pm 0.21}$ & 89.39$_{\pm 0.12}$ & 60.41$_{\pm 0.21}$ & 79.26$_{\pm 0.15}$ & 67.48$_{\pm 0.22}$ & 87.97$_{\pm 0.11}$ \\
  \hline
  FRN+TDM \cite{lee2022task} & CVPR 2022 & 72.01$_{\pm 0.22}$ & 89.05$_{\pm 0.12}$ & 51.57$_{\pm 0.23}$ & 75.25$_{\pm 0.16}$ & 65.67$_{\pm 0.22}$ & 86.44$_{\pm 0.12}$ \\
  \hline
  Bi-FRN \cite{wu2023bi} & AAAI 2023 & \underline{79.08}$_{\pm 0.20}$ & \underline{92.22}$_{\pm 0.10}$ & 64.74$_{\pm 0.22}$ & \underline{81.29}$_{\pm 0.14}$ & 75.74$_{\pm 0.20}$ & \underline{91.58}$_{\pm 0.09}$ \\ 
  \hline
  C2-Net \cite{ma2024cross} & AAAI 2024 & - & - & \underline{66.42}$_{\pm 0.50}$ & 81.23$_{\pm 0.34}$ & \textbf{81.29}$_{\pm 0.45}$ & 91.08$_{\pm 0.26}$ \\
  \hline
  \textbf{Our CausalFSFG} & - & \textbf{81.94$_{\pm 0.19}$} & \textbf{93.33$_{\pm 0.10}$} & \textbf{67.56$_{\pm 0.22}$} & \textbf{82.83$_{\pm 0.14}$} & \underline{79.96}$_{\pm 0.19}$ & \textbf{93.07$_{\pm 0.09}$} \\
  \hline
  \end{tabular}
  \end{center}
  \end{table*}
  
  \begin{table*}
  \begin{center}
  \caption{Comparison results mean$_{\pm \text{std}}$ on the CUB, Dogs, and Cars datasets with the ResNet-12 backbone. \textbf{Bold value} indicates the best performance and \underline{underline value} indicates the suboptimal performance.}
  \label{resnet12}
  \begin{tabular}{|c|c|c|c|c|c|c|c|}
  \hline
  \multirow{2}*{Method} & \multirow{2}*{Published In} & \multicolumn{2}{c|}{CUB} & \multicolumn{2}{c|}{Dogs} & \multicolumn{2}{c|}{Cars}\\ \cline{3-8}
  & & 1-shot & 5-shot & 1-shot & 5-shot & 1-shot & 5-shot \\
  \hline
  \hline
  ProtoNet \cite{snell2017prototypical} & NeurIPS 2017 & 81.02$_{\pm 0.20}$ & 91.93$_{\pm 0.11}$ & 73.81$_{\pm 0.21}$ & 87.39$_{\pm 0.12}$ & 85.46$_{\pm 0.19}$ & 95.08$_{\pm 0.08}$ \\
  \hline
  CTX \cite{doersch2020crosstransformers} & NeurIPS 2020 & 80.39$_{\pm 0.20}$ & 91.01$_{\pm 0.11}$ & 73.22$_{\pm 0.22}$ & 85.90$_{\pm 0.13}$ & 85.03$_{\pm 0.19}$ & 92.63$_{\pm 0.11}$ \\
  \hline
  DeepEMD \cite{zhang2020deepemd} & CVPR 2020 & 75.59$_{\pm 0.30}$ & 88.23$_{\pm 0.18}$ & 70.38$_{\pm 0.30}$ & 85.24$_{\pm 0.18}$ & 80.62$_{\pm 0.26}$ & 92.63$_{\pm 0.13}$ \\
  \hline
  FRN \cite{wertheimer2021few} & CVPR 2021 & 84.30$_{\pm 0.18}$ & 93.34$_{\pm 0.10}$ & 76.76$_{\pm 0.21}$ & 88.74$_{\pm 0.12}$ & 88.01$_{\pm 0.17}$ & 95.75$_{\pm 0.07}$ \\
  \hline
  FRN+TDM \cite{lee2022task} & CVPR 2022 & 85.15$_{\pm 0.18}$ & 93.99$_{\pm 0.09}$ & \underline{78.02}$_{\pm 0.20}$ & \underline{89.85}$_{\pm 0.11}$ & 88.92$_{\pm 0.16}$ & 96.88$_{\pm 0.06}$ \\
  \hline
  Bi-FRN \cite{wu2023bi} & AAAI 2023 & \underline{85.44}$_{\pm 0.18}$ & \underline{94.73}$_{\pm 0.09}$ & 76.89$_{\pm 0.21}$ & 88.27$_{\pm 0.12}$ & \underline{90.44}$_{\pm 0.15}$ & \underline{97.49}$_{\pm 0.05}$ \\
  \hline
  RAsD \cite{liu2024robust} & TMM 2024 & - & - & 73.75$_{\pm 0.93}$ & 86.65$_{\pm 0.54}$ & 87.27$_{\pm 0.70}$ & 95.01$_{\pm 0.49}$ \\
  \hline
  C2-Net \cite{ma2024cross} & AAAI 2024 & - & - & 75.50$_{\pm 0.49}$ & 87.65$_{\pm 0.28}$ & 88.96$_{\pm 0.37}$ & 95.16$_{\pm 0.20}$ \\
  \hline
  \textbf{Our CausalFSFG} & - & \textbf{87.05$_{\pm 0.17}$} & \textbf{95.26$_{\pm 0.08}$} & \textbf{78.79}$_{\pm 0.20}$ & \textbf{90.07}$_{\pm 0.11}$ & \textbf{90.71}$_{\pm 0.14}$ & \textbf{97.60}$_{\pm 0.05}$ \\
  \hline
  \end{tabular}
  \end{center}
  \end{table*}
  
\subsection{Experimental Results}
We compare the proposed CausalFSFG approach with state-of-the-art methods on the three fine-grained datasets with both meta-learning paradigms, and the results are shown in Table \ref{conv4}, \ref{resnet12}, \ref{meta-4}, and \ref{meta-12}. \par

\begin{itemize}
  \item Table \ref{conv4} presents the comparison results with the Conv-4 backbone following the meta-learning paradigm in \cite{wu2023bi}. The proposed CausalFSFG approach achieves superior performance on most experimental settings, achieving \textbf{81.94\%}, \textbf{93.33\%}, \textbf{67.56\%}, \textbf{82.83\%}, \textbf{79.96\%}, \textbf{93.07\%} for the 1-shot and 5-shot settings on the CUB, Dogs, and Cars datasets respectively. Compared with the representative SOTA methods BiFRN \cite{wu2023bi} which accommodates for inter-class and intra-class variations through a bi-directional feature reconstruction, our CauFSFG approach achieves \textbf{2.86\%}, \textbf{1.11\%}, \textbf{2.82\%}, \textbf{1.54\%}, \textbf{4.22\%}, \textbf{1.49\%} performance improvements respectively. We attribute the improvements brought by our CausalFSFG approach to the utilizing of the causal intervention to address the biased distribution of the limited support samples, which breaks the restriction of the selected samples and enables the model to learn a more general distribution of the entire fine-grained dataset.
  \item Table \ref{resnet12} presents the comparison results with the ResNet-12 backbone following the meta-learning paradigm in \cite{wu2023bi}. Notice that the ResNet-12 backbone is a more complicated and powerful network than the Conv-4 backbone, which naturally learns a better data distribution and weakens the effectiveness of our proposed approach to some extent, and leads to less significant performance gains compared with the Conv-4 backbone. In spite of this, the proposed CausalFSFG approach still achieves the best performance of \textbf{87.05\%}, \textbf{95.26\%}, \textbf{78.79\%}, \textbf{90.07\%}, \textbf{90.71\%}, \textbf{97.60\%} for the 1-shot and 5-shot settings on the CUB, Dogs, and Cars datasets respectively.
  \item Table \ref{meta-4} and \ref{meta-12} present the comparison results on the CUB dataset for both the Conv-4 and the ResNet-12 backbones following the meta-learning paradigm in \cite{lee2022task}. Our CausalFSFG approach maintains the best performance of \textbf{79.12\%}, \textbf{92.01\%}, \textbf{85.01\%}, \textbf{94.56\%} for the 1-shot and 5-shot settings, respectively. Compared with the representative SOTA methods LCCRN \cite{li2023locally} which learns local content-enriched features, our CausalFSFG approach also achieves \textbf{0.46\%}, \textbf{2.58\%}, \textbf{1.65\%}, \textbf{0.93\%} performance gains, which verifies the effectiveness of our CausalFSFG approach under different training paradigms.
\end{itemize}

\begin{table}
  \begin{center}
    \caption{Comparison results on the CUB dataset with the Conv-4 backbone. \textbf{Bold value} indicates the best performance and \underline{underline value} indicates the suboptimal performance.}
    \label{meta-4}
    \resizebox{1.0\linewidth}{!}{
    \begin{tabular}{|c|c|c|c|}
    \hline
    Method & Published In&  1-shot & 5-shot \\
    \hline
    \hline
    ProtoNet \cite{snell2017prototypical} & NeurIPS 2017 & 61.82$_{\pm 0.23}$ & 83.37$_{\pm 0.75}$  \\
    \hline
    FRN \cite{wertheimer2021few} & CVPR 2021 & 73.46$_{\pm 0.21}$ & 88.13$_{\pm 0.13}$ \\
    \hline
    Dual Att-Net \cite{xu2022dual} & AAAI 2022 & 72.89$_{\pm 0.50}$ & 86.60$_{\pm 0.31}$ \\
    \hline
     FRN+TDM\cite{lee2022task} & CVPR 2022 & 74.39$_{\pm 0.21}$ & 88.89$_{\pm 0.13}$ \\
    \hline
     LCCRN\cite{li2023locally} & TCSVT 2023 & 76.22$_{\pm 0.21}$ & 89.39$_{\pm 0.13}$ \\
     \hline
     RSaD\cite{liu2024robust} & TMM 2024 & 71.15$_{\pm 0.92}$ & 84.03$_{\pm 0.62}$ \\
    \hline
    FicNet\cite{zhu2024few} & TMM 2024 & 75.27$_{\pm 0.61}$ & 88.48$_{\pm 0.37}$ \\
    \hline
    C2-Net \cite{ma2024cross} & AAAI 2024 & \underline{78.66}$_{\pm 0.46}$ & \underline{89.43}$_{\pm 0.28}$ \\
    \hline
    \textbf{Our CausalFSFG} & - &  \textbf{79.12}$_{\pm 0.20}$ & \textbf{92.01}$_{\pm 0.11}$ \\
    \hline
    \end{tabular}}
  \end{center}
\end{table}

  \subsection{Ablation Study}
  To further demonstrate the effectiveness of the proposed CausalFSFG approach, we evaluate the key components in our CausalFSFG framework on the CUB dataset, adopting the Conv-4 and ResNet-12 backbones respectively. The results of ablation experiments are presented in Table \ref{ablation}. We can observe that:
    (1) Based on the Conv-4 backbone, IMSE aggregates multi-scale features to achieve the few-shot fine-grained classification accuracy of $77.13\%$ and $88.24\%$ for the 1-shot and 5-shot configurations, which outperforms the baseline method by margins of $12.31\%$ and $2.5\%$ and verifies the effectiveness of conducting sample-level intervention for learning less biased feature distribution. The same trend is observed on the ResNet-12 backbone, where IMSE achieves $4.6\%$ and $1.18\%$ performance improvements.
    (2) IMFR reconstructs the features of query images to achieve the few-shot fine-grained classification accuracy of $73.51\%$ and $88.75\%$ for the 1-shot and 5-shot configurations, which outperforms the baseline method by margins of $8.69\%$ and $3.01\%$ and verifies the effectiveness of conducting feature-level intervention for extracting more discriminative features. The same trend is observed on the ResNet-12 backbone, where our IMFR module achieves $3.44\%$ and $2.45\%$ performance improvements.
    (3) By combining the IMSE and IMFR modules, our CausalFSFG approach first learns less biased feature distributions and then extracts more discriminative features, which eliminates spurious correlations caused by the few-shot condition and further achieves performance improvements of $17.12\%$, $7.59\%$, $6.03\%$, $3.33\%$ on both backbones, respectively.

  \subsection{Parameter Analysis}
  To further investigate the effectiveness of our proposed approach, we conduct parameter experiments about the embedding dim $\Gamma$, and the top-$k$ threshold on the CUB dataset to explore different designs in our framework.

  \begin{table}
    \begin{center}
      \caption{Comparison results on the CUB dataset with the ResNet-12 backbone. \textbf{Bold value} indicates the best performance and \underline{underline value} indicates the suboptimal performance.}
      \label{meta-12}
      \resizebox{1.0\linewidth}{!}{
      \begin{tabular}{|c|c|c|c|}
      \hline
       Method & Published In &  1-shot & 5-shot \\
      \hline
      \hline
      ProtoNet \cite{snell2017prototypical} & NeurIPS 2017 & 79.64$_{\pm 0.20}$ & 91.15$_{\pm 0.11}$ \\
      \hline
      FRN \cite{wertheimer2021few} & CVPR 2021 & 83.11$_{\pm 0.19}$ & 92.49$_{\pm 0.11}$ \\
      \hline
       FRN+TDM\cite{lee2022task} & CVPR 2022 & \underline{83.36}$_{\pm 0.19}$ & 92.80$_{\pm 0.10}$ \\
      \hline
      LAGPF \cite{tang2022learning} & PR 2023 & 78.73$_{\pm 0.84}$ & 89.77$_{\pm 0.47}$ \\
      \hline
      BSFA \cite{zha2023boosting} & TCSVT 2023 & 82.27$_{\pm 0.46}$ & 90.76$_{\pm 0.26}$ \\
      \hline
       LCCRN\cite{li2023locally} & TCSVT 2023 & 82.97$_{\pm 0.19}$ & \underline{93.63}$_{\pm 0.10}$ \\
       \hline
       FicNet\cite{zhu2024few} & TMM 2024 & 80.97$_{\pm 0.57}$ & 93.17$_{\pm 0.32}$ \\
       \hline
       RSaD\cite{liu2024robust} & TMM 2024 & 82.45$_{\pm 0.79}$ & 92.02$_{\pm 0.44}$ \\
      \hline
      \textbf{Our CausalFSFG} & - & \textbf{85.01}$_{\pm 0.18}$ & \textbf{94.56}$_{\pm 0.09}$ \\
      \hline
      \end{tabular}}
    \end{center}
  \end{table}
  
  \begin{table}
  \begin{center}
  \caption{Ablation study on the CUB dataset of different components. \textbf{Bold value} indicates the best performance.}
  \label{ablation}
  \resizebox{1.0\linewidth}{!}{
  \begin{tabular}{|c|c|c|c|c|c|}
  \hline
  \multirow{2}*{IMSE} & \multirow{2}*{IMFR} & \multicolumn{2}{c|}{Conv-4} & \multicolumn{2}{c|}{ResNet-12} \\ \cline{3-6}
  & & 1-shot & 5-shot & 1-shot & 5-shot \\
  \hline
  \hline
  & & 64.82$_{\pm 0.23}$ & 85.74$_{\pm 0.14}$ & 81.02$_{\pm 0.20}$ & 91.93$_{\pm 0.11}$ \\
  \hline
  \checkmark & & 77.13$_{\pm 0.21}$ & 88.24$_{\pm 0.13}$ & 85.62$_{\pm 0.18}$ & 93.11$_{\pm 0.10}$ \\
  \hline
  & \checkmark & 73.51$_{\pm 0.21}$ & 88.75$_{\pm 0.12}$ & 84.46$_{\pm 0.18}$ & 94.38$_{\pm 0.09}$ \\
  \hline
  \checkmark & \checkmark & \textbf{81.94}$_{\pm 0.19}$ & \textbf{93.33}$_{\pm 0.10}$ & \textbf{87.05}$_{\pm 0.17}$ & \textbf{95.26}$_{\pm 0.08}$ \\
  \hline
  \end{tabular}}
  \end{center}
  \end{table}

  \begin{table}[t]
    \begin{center}
    \caption{Parameter analysis on the CUB dataset of different embedding dims. \textbf{Bold value} indicates the best performance. }
    \label{embed}
    \resizebox{1.0\linewidth}{!}{\begin{tabular}{|c|c|c|c|c|}
    \hline
     \multirow{2}*{$\Gamma$} & \multicolumn{2}{c|}{Conv-4} & \multicolumn{2}{|c|}{ResNet-12} \\ \cline{2-5}
    & 1-shot & 5-shot & 1-shot & 5-shot \\
    \hline
    \hline
    64 & 80.88$_{\pm 0.20}$ & 92.88$_{\pm 0.10}$ & 86.44$_{\pm 0.17}$ & 95.04$_{\pm 0.08}$ \\
    \hline
    128 & \textbf{81.94}$_{\pm 0.19}$ & \textbf{93.33}$_{\pm 0.10}$ & 86.22$_{\pm 0.17}$ & 95.02$_{\pm 0.08}$ \\
    \hline
    256 & 81.71$_{\pm 0.19}$ & 93.08$_{\pm 0.10}$ & \textbf{87.05}$_{\pm 0.17}$ & \textbf{95.26}$_{\pm 0.08}$ \\
    \hline
    \end{tabular}}
    \end{center}
    \end{table}

  \begin{figure*}[!t]
    \centering
    \includegraphics[width=0.95\textwidth]{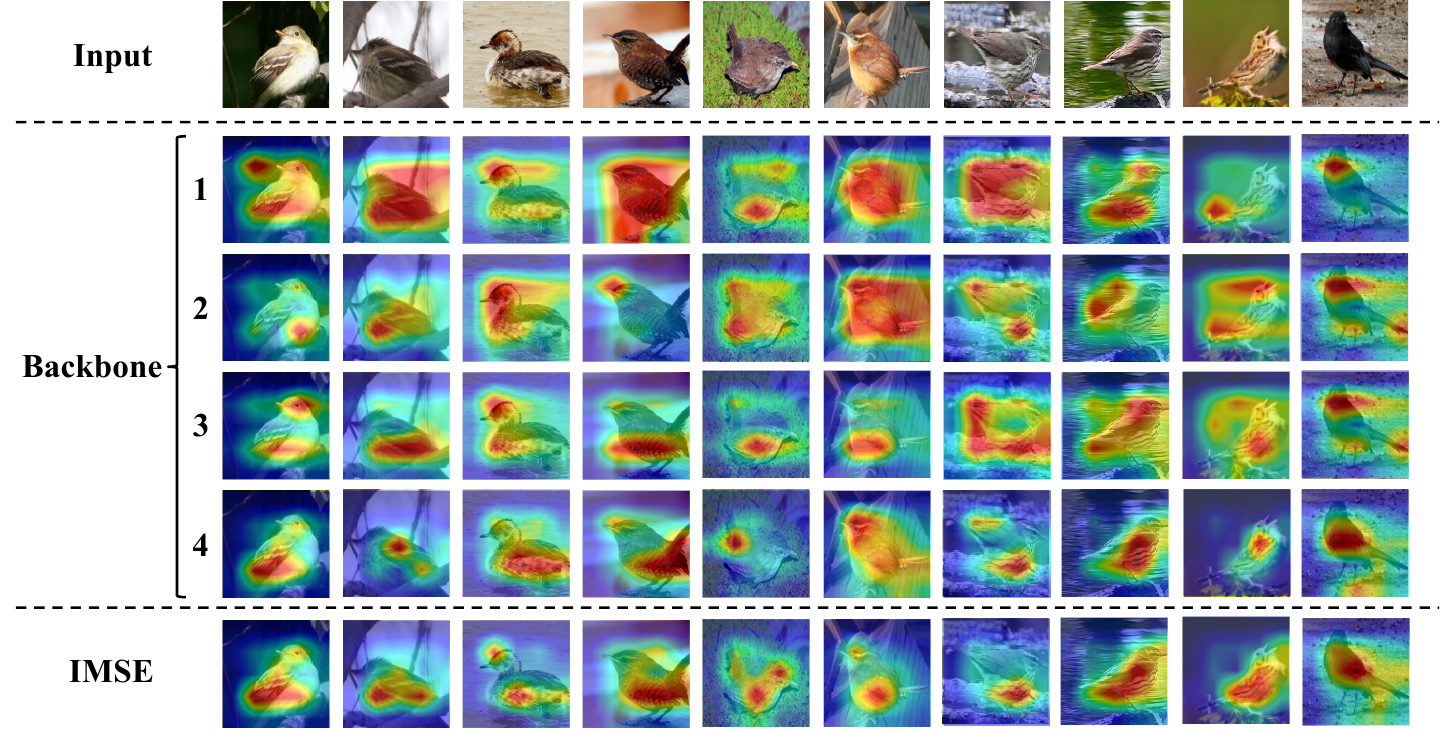}
    \caption{The visualization examples of the proposed CausalFSFG framework on the CUB dataset with the ResNet-12 backbone. The top block presents the input samples. The middle block presents the visualizations of the multi-scale features extracted by the first to fourth layers of the backbone, respectively. The bottom block presents the visualizations of the intervened feature generated by our IMSE module.}
    \label{visual}
  \end{figure*}
  
\paragraph{Embedding Dim}
The experimental results on the embedding dim are presented in Table \ref{embed}, where the value of embedding dim ranges in $\{62, 128, 256\}$. The best performances are achieved when the embedding dim is set as 128 and 256 for the Conv-4 and ResNet-12 backbones respectively. The different best values can be attributed to the different designs of the backbones. 
The Conv-4 backbone, with its straightforward architecture, generates  feature maps of size $64\times 5 \times 5$, but struggles with overfitting when the embedding dimension is raised to 256. In contrast, the ResNet-12 backbone, with its more intricate architecture, captures richer semantic information and generates larger feature maps of size $640\times 5 \times 5$. Consequently, classification accuracy using the ResNet-12 backbone continues to improve as the embedding dimension increases from 64 to 256.
  
  \begin{table}[h]
  \begin{center}
  \caption{Parameter analysis on the CUB dataset of different selective top-k numbers. \textbf{Bold value} indicates the best performance.}
  \label{top}
  \resizebox{1.0\linewidth}{!}{\begin{tabular}{|c|c|c|c|c|}
  \hline
   \multirow{2}*{$k$} & \multicolumn{2}{c|}{Conv-4} & \multicolumn{2}{c|}{ResNet-12} \\ \cline{2-5}
  & 1-shot & 5-shot & 1-shot & 5-shot \\
  \hline
  \hline
  3 & 81.77$_{\pm 0.19}$ & 93.23$_{\pm 0.10}$ & \textbf{87.05}$_{\pm 0.17}$ & 95.26$_{\pm 0.08}$ \\
  \hline
  5 & \textbf{81.94}$_{\pm 0.19}$ & \textbf{93.33}$_{\pm 0.10}$ & 86.83$_{\pm 0.17}$ & 95.25$_{\pm 0.08}$ \\
  \hline
  7 & 81.53$_{\pm 0.19}$ & 93.01$_{\pm 0.10}$ & 86.82$_{\pm 0.17}$ & 95.19$_{\pm 0.08}$ \\
  \hline
  10 & 81.71$_{\pm 0.19}$ & 93.23$_{\pm 0.10}$ & 86.97$_{\pm 0.17}$ & \textbf{95.33}$_{\pm 0.08}$ \\
  \hline  
  \end{tabular}}
  \end{center}
  \end{table}
  
\paragraph{Top-k Thresholds}
In this section, we investigate the effect of top-k thresholds on model performance, as shown in Table \ref{top}. We can observe that slight performance fluctuation happens with different thresholds, which indicates that our proposed approach is relatively insensitive to the value of thresholds. In particular, we set the value of thresholds as 5 with Conv-4 and 3 with ResNet-12 for the best performance.

\begin{figure}[!t]
    \centering
    \includegraphics[width=1.0\columnwidth]{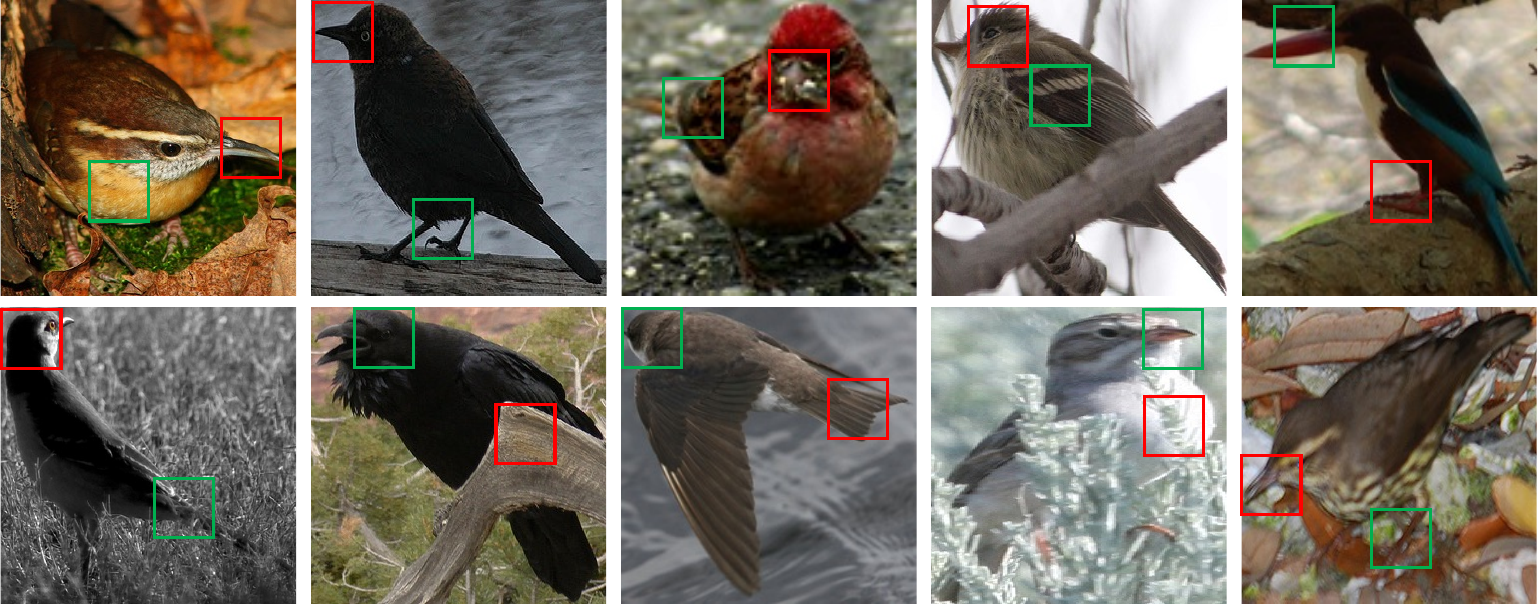}
    \caption{Visualizations of erroneous results on the CUB test set. Red boxes indicate false image contents identified by the model, and green boxes highlight discriminative image contents for correct classification.}
    \label{fail}
\end{figure}

\begin{table}
    \begin{center}
      \caption{Computational complexity comparison against Bi-FRN with the Conv-4 backbone.}
      \label{complexity}
      \resizebox{1.0\linewidth}{!}{
      \begin{tabular}{|c|c|c|c|}
      \hline
       Method & Params (K) & Memory (G) & GFLOPs \\
      \hline
      \hline
       Bi-FRN~\cite{wu2023bi} & 150.53 & 13.34 & 60.30 \\
      \hline
      \textbf{Our CausalFSFG} & 168.26 & 14.81 & 66.07 \\
      \hline
      \end{tabular}}
    \end{center}
\end{table}

\subsection{Visualization Experiments}
Fig. \ref{visual} illustrates the visualization results of the proposed CausalFSFG framework. Specifically, the input images are presented in the first row. The visualization results of the multi-scale features extracted from different layers are present in the second to fifth rows, and the visualization results of our intervened features are shown in the last row. It can be observed that: 
(1) The multi-scale features contain complementary information concerning each other, which focus on different parts of the input sample.
(2) The multi-scale features alone are affected by the biased distribution of the limited samples and fail to precisely locate the salient object and its discriminative parts. 
(3) Our intervened features alleviate the biased distribution and locate the salient objects and capture corresponding discriminative parts with higher precision.

\textbf{Failure case analysis.} Figure~\ref{fail} presents the visualization of erroneous results in Figure 5, where red boxes indicate false image contents identified by the model, and green boxes highlight discriminative image contents for correct classification. It can be observed that misclassifications often occur when discriminative features are visually compromised due to extreme poses and occlusions, or misled by salient yet spurious background cues when the foreground object is ambiguous.

\subsection{Complexity Analysis}
Table~\ref{complexity} compares the computational complexity of our CausalFSFG approach against the best comparison method Bi-FRN~\cite{wu2023bi}, with the Conv-4 backbone. It can be observed that our CausalFSFG achieves superior FS-FGVC performance with only slightly increased computational burden.
  
\section{Conclusion}
\label{conclusion}
In this paper, we propose a novel CausalFSFG approach for few-shot fine-grained visual categorization, which reformulates the FS-FGVC problem from the causal perspective to alleviate the biased data distribution inherent in the few-shot condition through causal intervention. We first align the FS-FGVC problem with a Structural Causal Model (SCM) assumption, where the few-shot condition and the inherent fine-grained nature collectively constitute an unobservable confounder, restricting the classification performance by introducing spurious correlations. To further mitigate this issue, an Interventional Multi-Scale Encoder (IMSE) module and an Interventional Masked Feature Reconstruction (IMFR) module are proposed to conduct the sample and feature level intervention respectively, which eliminates the spurious correlations and reveals real causalities from inputs to subcategories. Extensive experiments and analyses on three public fine-grained datasets validate the superiority and practicality of the proposed CausalFSFG approach.  \par 

In our ongoing efforts, we aim to enhance this research by consolidating sample and feature level interventions into a unified network architecture. This integration is expected to offer a more concise and efficient solution for addressing biased distributions and improving overall model efficacy.
 
\bibliographystyle{IEEEtran}
\bibliography{tmm24}

@article{tang2022learning,
  title={Learning Attention-Guided Pyramidal Features for Few-shot Fine-grained Recognition},
  author={Tang, Hao and Yuan, Chengcheng and Li, Zechao and Tang, Jinhui},
  journal={Pattern Recognition},
  pages={108792},
  year={2022},
  publisher={Elsevier}
}

@article{huang2020low,
  title={Low-rank pairwise alignment bilinear network for few-shot fine-grained image classification},
  author={Huang, Huaxi and Zhang, Junjie and Zhang, Jian and Xu, Jingsong and Wu, Qiang},
  journal={IEEE Transactions on Multimedia},
  volume={23},
  pages={1666--1680},
  year={2020},
  publisher={IEEE}
}

@inproceedings{li2019revisiting,
  title={Revisiting local descriptor based image-to-class measure for few-shot learning},
  author={Li, Wenbin and Wang, Lei and Xu, Jinglin and Huo, Jing and Gao, Yang and Luo, Jiebo},
  booktitle={Proceedings of the IEEE/CVF Conference on Computer Vision and Pattern Recognition},
  pages={7260--7268},
  year={2019}
}

@inproceedings{yoon2019tapnet,
  title={Tapnet: Neural network augmented with task-adaptive projection for few-shot learning},
  author={Yoon, Sung Whan and Seo, Jun and Moon, Jaekyun},
  booktitle={International Conference on Machine Learning},
  pages={7115--7123},
  year={2019},
  organization={PMLR}
}

@inproceedings{guo2020attentive,
  title={Attentive weights generation for few shot learning via information maximization},
  author={Guo, Yiluan and Cheung, Ngai-Man},
  booktitle={Proceedings of the IEEE/CVF Conference on Computer Vision and Pattern Recognition},
  pages={13499--13508},
  year={2020}
}

@inproceedings{yang2020dpgn,
  title={Dpgn: Distribution propagation graph network for few-shot learning},
  author={Yang, Ling and Li, Liangliang and Zhang, Zilun and Zhou, Xinyu and Zhou, Erjin and Liu, Yu},
  booktitle={Proceedings of the IEEE/CVF Conference on Computer Vision and Pattern Recognition},
  pages={13390--13399},
  year={2020}
}

@inproceedings{lee2022task,
  title={Task Discrepancy Maximization for Fine-Grained Few-Shot Classification},
  author={Lee, SuBeen and Moon, WonJun and Heo, Jae-Pil},
  booktitle={Proceedings of the IEEE/CVF Conference on Computer Vision and Pattern Recognition},
  pages={5331--5340},
  year={2022}
}

@article{vinyals2016matching,
  title={Matching networks for one shot learning},
  author={Vinyals, Oriol and Blundell, Charles and Lillicrap, Timothy and Wierstra, Daan and others},
  journal={Advances in neural information processing systems},
  volume={29},
  year={2016}
}

@article{li2020bsnet,
  title={BSNet: Bi-similarity network for few-shot fine-grained image classification},
  author={Li, Xiaoxu and Wu, Jijie and Sun, Zhuo and Ma, Zhanyu and Cao, Jie and Xue, Jing-Hao},
  journal={IEEE Transactions on Image Processing},
  volume={30},
  pages={1318--1331},
  year={2020},
  publisher={IEEE}
}

@inproceedings{sung2018learning,
  title={Learning to compare: Relation network for few-shot learning},
  author={Sung, Flood and Yang, Yongxin and Zhang, Li and Xiang, Tao and Torr, Philip HS and Hospedales, Timothy M},
  booktitle={Proceedings of the IEEE conference on computer vision and pattern recognition},
  pages={1199--1208},
  year={2018}
}

@inproceedings{lee2019meta,
  title={Meta-learning with differentiable convex optimization},
  author={Lee, Kwonjoon and Maji, Subhransu and Ravichandran, Avinash and Soatto, Stefano},
  booktitle={Proceedings of the IEEE/CVF Conference on Computer Vision and Pattern Recognition},
  pages={10657--10665},
  year={2019}
}

@inproceedings{lifchitz2019dense,
  title={Dense classification and implanting for few-shot learning},
  author={Lifchitz, Yann and Avrithis, Yannis and Picard, Sylvaine and Bursuc, Andrei},
  booktitle={Proceedings of the IEEE/CVF Conference on Computer Vision and Pattern Recognition},
  pages={9258--9267},
  year={2019}
}

@inproceedings{jamal2019task,
  title={Task agnostic meta-learning for few-shot learning},
  author={Jamal, Muhammad Abdullah and Qi, Guo-Jun},
  booktitle={Proceedings of the IEEE/CVF Conference on Computer Vision and Pattern Recognition},
  pages={11719--11727},
  year={2019}
}

@article{antoniou2018train,
  title={How to train your MAML},
  author={Antoniou, Antreas and Edwards, Harrison and Storkey, Amos},
  journal={arXiv preprint arXiv:1810.09502},
  year={2018}
}

@inproceedings{finn2017model,
  title={Model-agnostic meta-learning for fast adaptation of deep networks},
  author={Finn, Chelsea and Abbeel, Pieter and Levine, Sergey},
  booktitle={International conference on machine learning},
  pages={1126--1135},
  year={2017},
  organization={PMLR}
}

@inproceedings{zhu2020multi,
  title={Multi-attention Meta Learning for Few-shot Fine-grained Image Recognition.},
  author={Zhu, Yaohui and Liu, Chenlong and Jiang, Shuqiang},
  booktitle={IJCAI},
  pages={1090--1096},
  year={2020}
}

@article{song2020bi,
  title={Bi-modal progressive mask attention for fine-grained recognition},
  author={Song, Kaitao and Wei, Xiu-Shen and Shu, Xiangbo and Song, Ren-Jie and Lu, Jianfeng},
  journal={IEEE Transactions on Image Processing},
  volume={29},
  pages={7006--7018},
  year={2020},
  publisher={IEEE}
}

@inproceedings{zhang2020web,
  title={Web-supervised network with softly update-drop training for fine-grained visual classification},
  author={Zhang, Chuanyi and Yao, Yazhou and Liu, Huafeng and Xie, Guo-Sen and Shu, Xiangbo and Zhou, Tianfei and Zhang, Zheng and Shen, Fumin and Tang, Zhenmin},
  booktitle={Proceedings of the AAAI Conference on Artificial Intelligence},
  volume={34},
  number={07},
  pages={12781--12788},
  year={2020}
}

@inproceedings{sun2022sim,
  title={Sim-trans: Structure information modeling transformer for fine-grained visual categorization},
  author={Sun, Hongbo and He, Xiangteng and Peng, Yuxin},
  booktitle={Proceedings of the 30th ACM International Conference on Multimedia},
  pages={5853--5861},
  year={2022}
}

@article{he2019and,
  title={Which and how many regions to gaze: Focus discriminative regions for fine-grained visual categorization},
  author={He, Xiangteng and Peng, Yuxin and Zhao, Junjie},
  journal={International Journal of Computer Vision},
  volume={127},
  number={9},
  pages={1235--1255},
  year={2019},
  publisher={Springer}
}

@article{he2018fast,
  title={Fast fine-grained image classification via weakly supervised discriminative localization},
  author={He, Xiangteng and Peng, Yuxin and Zhao, Junjie},
  journal={IEEE Transactions on Circuits and Systems for Video Technology},
  volume={29},
  number={5},
  pages={1394--1407},
  year={2018},
  publisher={IEEE}
}

@article{snell2017prototypical,
  title={Prototypical networks for few-shot learning},
  author={Snell, Jake and Swersky, Kevin and Zemel, Richard},
  journal={Advances in neural information processing systems},
  volume={30},
  year={2017}
}

@article{wei2021fine,
  title={Fine-grained image analysis with deep learning: A survey},
  author={Wei, Xiu-Shen and Song, Yi-Zhe and Mac Aodha, Oisin and Wu, Jianxin and Peng, Yuxin and Tang, Jinhui and Yang, Jian and Belongie, Serge},
  journal={IEEE Transactions on Pattern Analysis and Machine Intelligence},
  year={2021},
  publisher={IEEE}
}

@article{xu2022dual,
  title={Dual Attention Networks for Few-Shot Fine-Grained Recognition},
  author={Xu, Shu-Lin and Zhang, Faen and Wei, Xiu-Shen and Wang, Jianhua},
  year={2022}
}

@inproceedings{krause20133d,
  title={3d object representations for fine-grained categorization},
  author={Krause, Jonathan and Stark, Michael and Deng, Jia and Fei-Fei, Li},
  booktitle={Proceedings of the IEEE international conference on computer vision workshops},
  pages={554--561},
  year={2013}
}

@inproceedings{khosla2011novel,
  title={Novel dataset for fine-grained image categorization: Stanford dogs},
  author={Khosla, Aditya and Jayadevaprakash, Nityananda and Yao, Bangpeng and Li, Fei-Fei},
  booktitle={Proc. CVPR workshop on fine-grained visual categorization (FGVC)},
  volume={2},
  number={1},
  year={2011},
  organization={Citeseer}
}

@article{wah2011caltech,
  title={The caltech-ucsd birds-200-2011 dataset},
  author={Wah, Catherine and Branson, Steve and Welinder, Peter and Perona, Pietro and Belongie, Serge},
  year={2011},
  publisher={California Institute of Technology}
}

@article{yue2020interventional,
  title={Interventional few-shot learning},
  author={Yue, Zhongqi and Zhang, Hanwang and Sun, Qianru and Hua, Xian-Sheng},
  journal={Advances in neural information processing systems},
  volume={33},
  pages={2734--2746},
  year={2020}
}

@book{pearl2016causal,
  title={Causal inference in statistics: A primer},
  author={Pearl, Judea and Glymour, Madelyn and Jewell, Nicholas P},
  year={2016},
  publisher={John Wiley \& Sons}
}

@article{pearl2000models,
  title={Models, reasoning and inference},
  author={Pearl, Judea and others},
  journal={Cambridge, UK: CambridgeUniversityPress},
  volume={19},
  number={2},
  pages={3},
  year={2000}
}

@inproceedings{wu2023bi,
  title={Bi-directional feature reconstruction network for fine-grained few-shot image classification},
  author={Wu, Jijie and Chang, Dongliang and Sain, Aneeshan and Li, Xiaoxu and Ma, Zhanyu and Cao, Jie and Guo, Jun and Song, Yi-Zhe},
  booktitle={Proceedings of the AAAI Conference on Artificial Intelligence},
  volume={37},
  number={3},
  pages={2821--2829},
  year={2023}
}

@inproceedings{wu2019parn,
  title={Parn: Position-aware relation networks for few-shot learning},
  author={Wu, Ziyang and Li, Yuwei and Guo, Lihua and Jia, Kui},
  booktitle={Proceedings of the IEEE/CVF international conference on computer vision},
  pages={6659--6667},
  year={2019}
}

@inproceedings{hao2019collect,
  title={Collect and select: Semantic alignment metric learning for few-shot learning},
  author={Hao, Fusheng and He, Fengxiang and Cheng, Jun and Wang, Lei and Cao, Jianzhong and Tao, Dacheng},
  booktitle={Proceedings of the IEEE/CVF international Conference on Computer Vision},
  pages={8460--8469},
  year={2019}
}

@inproceedings{zhang2020deepemd,
  title={Deepemd: Few-shot image classification with differentiable earth mover's distance and structured classifiers},
  author={Zhang, Chi and Cai, Yujun and Lin, Guosheng and Shen, Chunhua},
  booktitle={Proceedings of the IEEE/CVF conference on computer vision and pattern recognition},
  pages={12203--12213},
  year={2020}
}

@article{doersch2020crosstransformers,
  title={Crosstransformers: spatially-aware few-shot transfer},
  author={Doersch, Carl and Gupta, Ankush and Zisserman, Andrew},
  journal={Advances in Neural Information Processing Systems},
  volume={33},
  pages={21981--21993},
  year={2020}
}

@inproceedings{wertheimer2021few,
  title={Few-shot classification with feature map reconstruction networks},
  author={Wertheimer, Davis and Tang, Luming and Hariharan, Bharath},
  booktitle={Proceedings of the IEEE/CVF conference on computer vision and pattern recognition},
  pages={8012--8021},
  year={2021}
}

@article{magliacane2018domain,
  title={Domain adaptation by using causal inference to predict invariant conditional distributions},
  author={Magliacane, Sara and Van Ommen, Thijs and Claassen, Tom and Bongers, Stephan and Versteeg, Philip and Mooij, Joris M},
  journal={Advances in neural information processing systems},
  volume={31},
  year={2018}
}

@article{bengio2019meta,
  title={A meta-transfer objective for learning to disentangle causal mechanisms},
  author={Bengio, Yoshua and Deleu, Tristan and Rahaman, Nasim and Ke, Rosemary and Lachapelle, S{\'e}bastien and Bilaniuk, Olexa and Goyal, Anirudh and Pal, Christopher},
  journal={arXiv preprint arXiv:1901.10912},
  year={2019}
}

@article{tang2020long,
  title={Long-tailed classification by keeping the good and removing the bad momentum causal effect},
  author={Tang, Kaihua and Huang, Jianqiang and Zhang, Hanwang},
  journal={Advances in Neural Information Processing Systems},
  volume={33},
  pages={1513--1524},
  year={2020}
}

@article{zhang2020causal,
  title={Causal intervention for weakly-supervised semantic segmentation},
  author={Zhang, Dong and Zhang, Hanwang and Tang, Jinhui and Hua, Xian-Sheng and Sun, Qianru},
  journal={Advances in Neural Information Processing Systems},
  volume={33},
  pages={655--666},
  year={2020}
}

@article{chalupka2014visual,
  title={Visual causal feature learning},
  author={Chalupka, Krzysztof and Perona, Pietro and Eberhardt, Frederick},
  journal={arXiv preprint arXiv:1412.2309},
  year={2014}
}

@inproceedings{du2020fine,
  title={Fine-grained visual classification via progressive multi-granularity training of jigsaw patches},
  author={Du, Ruoyi and Chang, Dongliang and Bhunia, Ayan Kumar and Xie, Jiyang and Ma, Zhanyu and Song, Yi-Zhe and Guo, Jun},
  booktitle={European Conference on Computer Vision},
  pages={153--168},
  year={2020},
  organization={Springer}
}

@inproceedings{lin2017feature,
  title={Feature pyramid networks for object detection},
  author={Lin, Tsung-Yi and Doll{\'a}r, Piotr and Girshick, Ross and He, Kaiming and Hariharan, Bharath and Belongie, Serge},
  booktitle={Proceedings of the IEEE conference on computer vision and pattern recognition},
  pages={2117--2125},
  year={2017}
}

@article{li2023locally,
  title={Locally-Enriched Cross-Reconstruction for Few-Shot Fine-Grained Image Classification},
  author={Li, Xiaoxu and Song, Qi and Wu, Jijie and Zhu, Rui and Ma, Zhanyu and Xue, Jing-Hao},
  journal={IEEE Transactions on Circuits and Systems for Video Technology},
  year={2023},
  publisher={IEEE}
}

@article{zha2023boosting,
  title={Boosting few-shot fine-grained recognition with background suppression and foreground alignment},
  author={Zha, Zican and Tang, Hao and Sun, Yunlian and Tang, Jinhui},
  journal={IEEE Transactions on Circuits and Systems for Video Technology},
  year={2023},
  publisher={IEEE}
}

@inproceedings{he2018only,
  title={Only learn one sample: Fine-grained visual categorization with one sample training},
  author={He, Xiangteng and Peng, Yuxin},
  booktitle={Proceedings of the 26th ACM international conference on Multimedia},
  pages={1372--1380},
  year={2018}
}

@article{peng2017object,
  title={Object-part attention model for fine-grained image classification},
  author={Peng, Yuxin and He, Xiangteng and Zhao, Junjie},
  journal={IEEE Transactions on Image Processing},
  volume={27},
  number={3},
  pages={1487--1500},
  year={2017},
  publisher={IEEE}
}

@article{liu2023transifc,
  title={TransIFC: invariant cues-aware feature concentration learning for efficient fine-grained bird image classification},
  author={Liu, Hai and Zhang, Cheng and Deng, Yongjian and Xie, Bochen and Liu, Tingting and Zhang, Zhaoli and Li, You-Fu},
  journal={IEEE Transactions on Multimedia},
  year={2023},
  publisher={IEEE}
}

@article{wang2023learning,
  title={Learning Mutually Exclusive Part Representations for Fine-grained Image Classification},
  author={Wang, Chuanming and Fu, Huiyuan and Ma, Huadong},
  journal={IEEE Transactions on Multimedia},
  year={2023},
  publisher={IEEE}
}

@article{zhang2022fine,
  title={Fine-Grained Image Classification by Class and Image-Specific Decomposition with Multiple Views},
  author={Zhang, Chunjie and Bai, Huihui and Zhao, Yao},
  journal={IEEE Transactions on Multimedia},
  year={2022},
  publisher={IEEE}
}

@article{sun2023hcl,
  title={HCL: Hierarchical Consistency Learning for Webly Supervised Fine-Grained Recognition},
  author={Sun, Hongbo and He, Xiangteng and Peng, Yuxin},
  journal={IEEE Transactions on Multimedia},
  year={2023},
  publisher={IEEE}
}

@article{lake2015human,
  title={Human-level concept learning through probabilistic program induction},
  author={Lake, Brenden M and Salakhutdinov, Ruslan and Tenenbaum, Joshua B},
  journal={Science},
  volume={350},
  number={6266},
  pages={1332--1338},
  year={2015},
  publisher={American Association for the Advancement of Science}
}

@article{liu2024robust,
  title={Robust Saliency-Aware Distillation for Few-shot Fine-grained Visual Recognition},
  author={Liu, Haiqi and Chen, CL Philip and Gong, Xinrong and Zhang, Tong},
  journal={IEEE Transactions on Multimedia},
  year={2024},
  publisher={IEEE}
}

@inproceedings{ma2024cross,
  title={Cross-Layer and Cross-Sample Feature Optimization Network for Few-Shot Fine-Grained Image Classification},
  author={Ma, Zhen-Xiang and Chen, Zhen-Duo and Zhao, Li-Jun and Zhang, Zi-Chao and Luo, Xin and Xu, Xin-Shun},
  booktitle={Proceedings of the AAAI Conference on Artificial Intelligence},
  volume={38},
  number={5},
  pages={4136--4144},
  year={2024}
}

@article{zhu2024few,
  title={Few-shot fine-grained image classification via multi-frequency neighborhood and double-cross modulation},
  author={Zhu, Hegui and Gao, Zhan and Wang, Jiayi and Zhou, Yange and Li, Chengqing},
  journal={IEEE Transactions on Multimedia},
  year={2024},
  publisher={IEEE}
}

@inproceedings{huang2021attributes,
  title={Attributes-guided and pure-visual attention alignment for few-shot recognition},
  author={Huang, Siteng and Zhang, Min and Kang, Yachen and Wang, Donglin},
  booktitle={Proceedings of the AAAI conference on artificial intelligence},
  volume={35},
  number={9},
  pages={7840--7847},
  year={2021}
}

@inproceedings{luo2021few,
  title={Few-shot learning via feature hallucination with variational inference},
  author={Luo, Qinxuan and Wang, Lingfeng and Lv, Jingguo and Xiang, Shiming and Pan, Chunhong},
  booktitle={Proceedings of the IEEE/CVF winter conference on applications of computer vision},
  pages={3963--3972},
  year={2021}
}

@article{wang2021fine,
  title={Fine-grained few shot learning with foreground object transformation},
  author={Wang, Chaofei and Song, Shiji and Yang, Qisen and Li, Xiang and Huang, Gao},
  journal={Neurocomputing},
  volume={466},
  pages={16--26},
  year={2021},
  publisher={Elsevier}
}

@article{sun2023t2l,
  title={T2L: Trans-transfer Learning for few-shot fine-grained visual categorization with extended adaptation},
  author={Sun, Nan and Yang, Po},
  journal={Knowledge-Based Systems},
  volume={264},
  pages={110329},
  year={2023},
  publisher={Elsevier}
}

@article{tam2025transfer,
  title={Transfer learning and mixup for fine-grained few-shot fungi classification},
  author={Tam, Jason Kahei and Gustineli, Murilo and Miyaguchi, Anthony},
  journal={arXiv preprint arXiv:2507.08248},
  year={2025}
}

@article{li2023attention,
  title={Attention-based deep meta-transfer learning for few-shot fine-grained fault diagnosis},
  author={Li, Chuanjiang and Li, Shaobo and Wang, Huan and Gu, Fengshou and Ball, Andrew D},
  journal={Knowledge-Based Systems},
  volume={264},
  pages={110345},
  year={2023},
  publisher={Elsevier}
}

@inproceedings{wu2021selective,
  title={Selective, structural, subtle: Trilinear spatial-awareness for few-shot fine-grained visual recognition},
  author={Wu, Heng and Zhao, Yifan and Li, Jia},
  booktitle={2021 IEEE International Conference on Multimedia and Expo (ICME)},
  pages={1--6},
  year={2021},
  organization={IEEE Computer Society}
}

@article{liu2023bilaterally,
  title={Bilaterally normalized scale-consistent sinkhorn distance for few-shot image classification},
  author={Liu, Yanbin and Zhu, Linchao and Wang, Xiaohan and Yamada, Makoto and Yang, Yi},
  journal={IEEE Transactions on Neural Networks and Learning Systems},
  volume={35},
  number={8},
  pages={11475--11485},
  year={2023},
  publisher={IEEE}
}

@article{wang2024multi,
  title={Multi-granularity part sampling attention for fine-grained visual classification},
  author={Wang, Jiahui and Xu, Qin and Jiang, Bo and Luo, Bin and Tang, Jinhui},
  journal={IEEE Transactions on Image Processing},
  year={2024},
  publisher={IEEE}
}

@inproceedings{jiang2024delving,
  title={Delving into multimodal prompting for fine-grained visual classification},
  author={Jiang, Xin and Tang, Hao and Gao, Junyao and Du, Xiaoyu and He, Shengfeng and Li, Zechao},
  booktitle={Proceedings of the AAAI conference on artificial intelligence},
  volume={38},
  number={3},
  pages={2570--2578},
  year={2024}
}

@inproceedings{cheng2024multi,
  title={Multi-modal Knowledge-Enhanced Fine-Grained Image Classification},
  author={Cheng, Suyan and Zhang, Feifei and Zhou, Haoliang and Xu, Changsheng},
  booktitle={Chinese Conference on Pattern Recognition and Computer Vision (PRCV)},
  pages={333--346},
  year={2024},
  organization={Springer}
}

@inproceedings{ma2024bi,
  title={Bi-directional Task-Guided Network for Few-Shot Fine-Grained Image Classification},
  author={Ma, Zhen-Xiang and Chen, Zhen-Duo and Zhao, Li-Jun and Zhang, Zi-Chao and Zheng, Tai and Luo, Xin and Xu, Xin-Shun},
  booktitle={Proceedings of the 32nd ACM International Conference on Multimedia},
  pages={8277--8286},
  year={2024}
}

@article{yu2024adaptive,
  title={Adaptive task-aware refining network for few-shot fine-grained image classification},
  author={Yu, Liyun and Guan, Ziyu and Zhao, Wei and Yang, Yaming and Tan, Jiale},
  journal={IEEE Transactions on Circuits and Systems for Video Technology},
  year={2024},
  publisher={IEEE}
}

@inproceedings{sun2025multi,
  title={Multi-Scale Cross-Modal Collaborative Reconstruction Network},
  author={Sun, Jie and Chen, Taotao and Luo, Tengxiang and Gu, Huamao and Tian, Yan and Li, Xiaoyu and Zheng, Jun and Fu, Jianhai},
  booktitle={2025 5th International Conference on Computer Vision, Application and Algorithm (CVAA)},
  pages={171--179},
  year={2025},
  organization={IEEE}
}

@article{wang2024unbiased,
  title={An unbiased feature estimation network for few-shot fine-grained image classification},
  author={Wang, Jiale and Lu, Jin and Yang, Junpo and Wang, Meijia and Zhang, Weichuan},
  journal={Sensors},
  volume={24},
  number={23},
  pages={7737},
  year={2024},
  publisher={MDPI}
}

@article{zhao2024angular,
  title={Angular isotonic loss guided multi-layer integration for few-shot fine-grained image classification},
  author={Zhao, Li-Jun and Chen, Zhen-Duo and Ma, Zhen-Xiang and Luo, Xin and Xu, Xin-Shun},
  journal={IEEE Transactions on Image Processing},
  volume={33},
  pages={3778--3792},
  year={2024},
  publisher={IEEE}
}

@article{wu2024bi,
  title={Bi-directional ensemble feature reconstruction network for few-shot fine-grained classification},
  author={Wu, Jijie and Chang, Dongliang and Sain, Aneeshan and Li, Xiaoxu and Ma, Zhanyu and Cao, Jie and Guo, Jun and Song, Yi-Zhe},
  journal={IEEE transactions on pattern analysis and machine intelligence},
  volume={46},
  number={9},
  pages={6082--6096},
  year={2024},
  publisher={IEEE}
}

@inproceedings{yang2024channel,
  title={Channel-Spatial Support-Query Cross-Attention for Fine-Grained Few-Shot Image Classification},
  author={Yang, Shicheng and Li, Xiaoxu and Chang, Dongliang and Ma, Zhanyu and Xue, Jing-Hao},
  booktitle={Proceedings of the 32nd ACM International Conference on Multimedia},
  pages={9175--9183},
  year={2024}
}

@article{wang2025interventional,
  title={Interventional Feature Generation for Few-shot Learning},
  author={Wang, Shuo and Lu, Jinda and Ben, Huixia and Hao, Yanbin and Gao, Xingyu and Wang, Meng},
  journal={ACM Transactions on Multimedia Computing, Communications and Applications},
  year={2025},
  publisher={ACM New York, NY}
}

\end{document}